\pdfoutput=1

\documentclass[11pt]{article}

\usepackage[preprint]{acl}

\usepackage{makecell}
\usepackage{multirow}
\usepackage{adjustbox}
\usepackage{amsmath}

\usepackage{times}
\usepackage{latexsym}

\usepackage{xcolor}
\usepackage{color}
\usepackage{colortbl}
\usepackage{bbding}
\usepackage{pifont}

\usepackage[T1]{fontenc}

\usepackage[utf8]{inputenc}

\usepackage{microtype}

\usepackage{inconsolata}

\usepackage{graphicx}
\usepackage{float}
\usepackage{stfloats}

\usepackage{longtable}  
\usepackage{geometry}   
\usepackage{array}
\usepackage{booktabs} 

\title{Reasoning is All You Need for Video Generalization: A Counterfactual Benchmark with Sub-question Evaluation}

\author{
    {\bf Qiji Zhou\footnotemark[1]}\textsuperscript{ \ }$^1$, {\bf YiFan Gong\footnotemark[1]}\textsuperscript{ \ }$ ^2$, {\bf Guangsheng Bao}$^{1}$, {\bf Hongjie Qiu}$^{2}$,\\ {\bf Jinqiang Li}$^{2}$
{\bf Xiangrong Zhu}$^{2}$, {\bf Huajian Zhang}$^{1}$,  {\bf Yue Zhang\footnotemark[2]}\textsuperscript{ \ }$^1$\\
$^{1}$School of Engineering, Westlake University\\
$^{2}$College of Computer Science and Technology, Hangzhou Dianzi University \\
\texttt{\{zhouqiji, baoguangsheng, zhanghuajian, zhangyue\}@westlake.edu.cn} \\
\texttt{\{gongyifan, qiuhongjie, lijinqiang, zhuxiangrong\}@hdu.edu.cn}
\\
}

\begin{document}
\maketitle
{
\renewcommand{\thefootnote}{\fnsymbol{footnote}}
\footnotetext[1]{Equal contribution.}
\footnotetext[2]{Corresponding Author.}
}

\begin{abstract}
Counterfactual reasoning is crucial for robust video understanding but remains underexplored in existing multimodal benchmarks. In this paper, we introduce \textbf{COVER} (\textbf{\underline{CO}}unterfactual \textbf{\underline{V}}id\textbf{\underline{E}}o \textbf{\underline{R}}easoning), a multidimensional multimodal benchmark that systematically evaluates MLLMs across the abstract-concrete and perception-cognition dimensions. 
Beyond prior multimodal benchmarks, COVER decomposes complex queries into structured sub-questions, enabling fine-grained reasoning analysis. Experiments on commercial and open-source models reveal a strong correlation between sub-question accuracy and counterfactual reasoning performance, highlighting the role of structured inference in video understanding. Furthermore, our results suggest a key insight: enhancing the reasoning capability of models is essential for improving the robustness of video understanding. COVER establishes a new standard for assessing MLLMs' logical reasoning abilities in dynamic environments. Our work is available at \href{https://github.com/gongyifan-hash/COVER-Benchmark}{https://github.com/gongyifan-hash/COVER-Benchmark}.
\end{abstract}

\section{Introduction}

\begin{figure*}[!htbp]
\centering
  \includegraphics[width=0.9\textwidth]{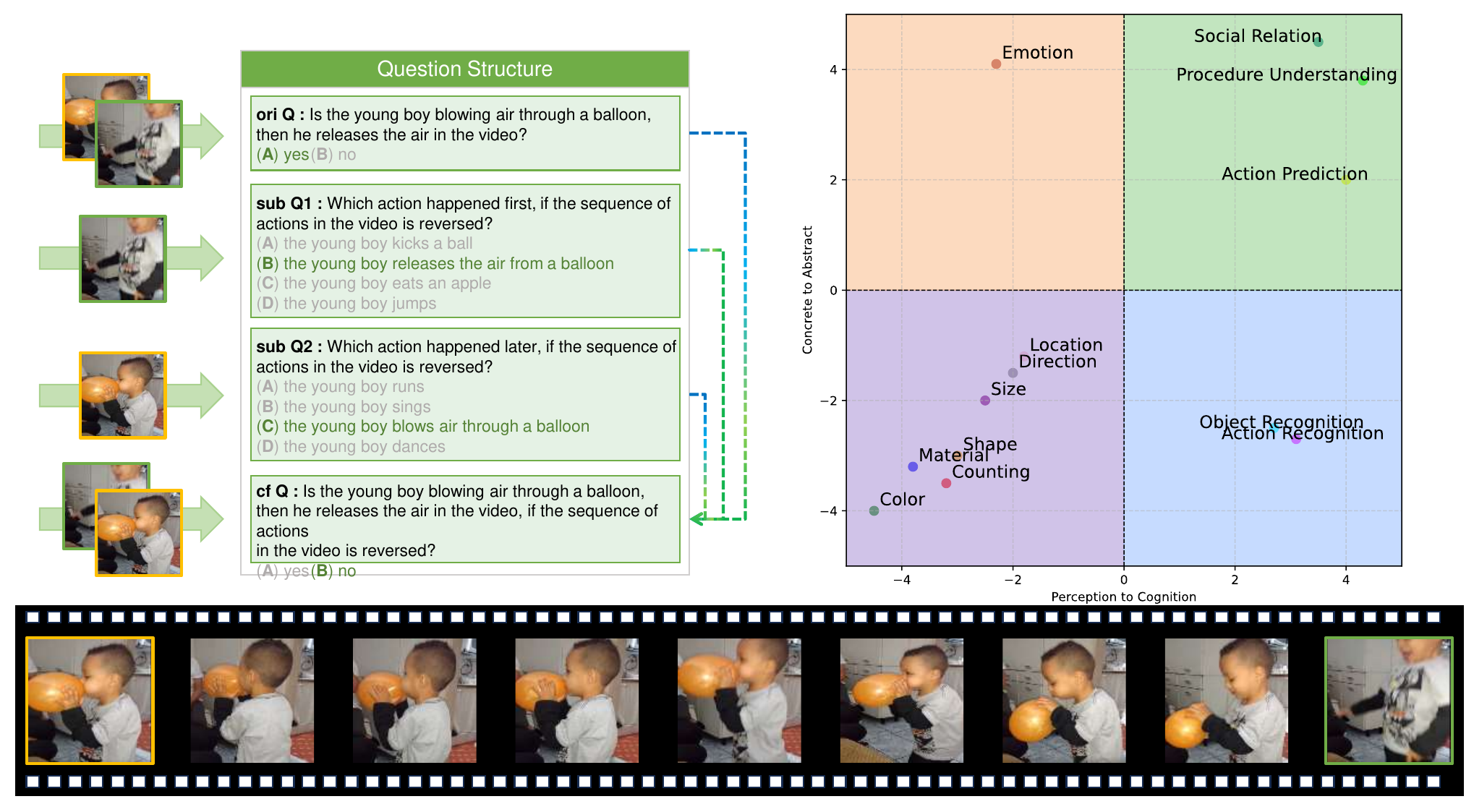}
  \caption{An example from the COVER benchmark. The ground-truth answers are highlighted in green. All data—including original questions, counterfactual questions, sub-questions, and videos—have been manually verified as part of COVER. The diagram in the upper right corner illustrates the division of each COVER task into four quadrants.}
  \label{fig:intro}
\end{figure*}


In recent years, the rapid advancement of large language models (LLMs) has spurred growing interest in multimodal large language models (MLLMs)~\cite{gpt4o,claude35,internvl2.5,llava-video,videollama3,qwen2-vl,vila-u}.
Various early benchmarks have been proposed to assess multimodal understanding ability of MLLMs, particularly in static images~\cite{mme,gqa,mmbench,mm-vet}.
More recently, benchmarks involving complex images and dynamic videos have emerged to evaluate MLLM's capabilities in temporal reasoning, spatio-temporal recognition, and object detection~\cite{videomme,Mvbench,seed-bench}.
Despite these advances, these benchmarks often overlook counterfactual reasoning, which is a critical component for evaluating inference in complex and realistic environments. 
As a result, they fall short of providing a comprehensive assessment of MLLMs' reasoning capabilities.

Counterfactual reasoning, which posits hypothetical alternatives to observed realities, is pivotal for advanced video inference and is closely tied to out-of-distribution generalization~\citep{yang-etal-2023-glue-x,bao-etal-2025-likely}. Previous work has attempted to construct counterfactual queries for images and videos~\citep{eyescandecive, cfmm, vitatecs, cripp, acquired}.
Most existing multimodal counterfactual benchmarks 
tend to focus on assessing subtask-specific robustness of reasoning ability~\citep{cfmm, ror, acquired}.
However, they do not assess the underlying factors that contribute to the robustness of these reasoning capabilities.
Such benchmarks often lack a systematic progression from abstract to concrete dimensions and from low-level perception to high-level cognition, limiting their ability to comprehensively capture multimodal reasoning processes in MLLMs. 
Furthermore, these benchmarks rarely investigate how robust video understanding interacts with stepwise reasoning in dynamic environments, leaving a gap in our assessment of advanced inference skills. 

To bridge this gap, we propose \textbf{COVER}, a counterfactual video reasoning benchmark driven by a multidimensional abstraction level evaluation mechanism. Unlike existing multimodal counterfactual benchmarks, which often focus on multitask-oriented questions, COVER systematically classifies tasks into four quadrants.
We define specific tasks for each quadrant to evaluate MLLMs’ diverse reasoning capabilities in complex video scenarios.
Beyond merely posing counterfactual questions, COVER introduces a \emph{sub-question} reasoning mechanism derived from necessary conditions, enabling a deeper evaluation of performance across MLLMs. This approach allows us to establish a connection between the accuracy of intermediate steps and the overall robustness of counterfactual reasoning. 
As shown in Figure~\ref{fig:intro}, when asked to determine whether a boy completes a series of actions in a specified order, COVER decomposes the problem into multiple steps, each representing a necessary condition. For instance, sub-question \emph{Q1} may inquire about which action occurs first in the reversed video, while sub-question \emph{Q2} targets the final action. This structured approach not only helps evaluate how a model adapts to event-sequence changes but also reveals its strengths and weaknesses in extracting and synthesizing critical information under counterfactual assumptions. By encompassing a broad range of abstraction levels, COVER stands as the most comprehensive dataset of its kind, paving the way for more rigorous and holistic evaluations of MLLMs’ dynamic and counterfactual reasoning capabilities.

Building on the COVER benchmark, we conducted a series of systematic experiments using both open-source and commercial closed-source models of varying scales.
Our results indicate a strong positive correlation between the models' sub-question accuracy and performance in counterfactual reasoning and robust video understanding.
The findings underscore the tight linkage between sophisticated inference capabilities and high-level video comprehension. 
Furthermore, we examine how automatically generated versus human-guided sub-question decomposition (chain-of-thought, CoT \cite{CoT})  influences complex reasoning and identifies the key factors impacting inference accuracy in MLLMs. 
Through these experiments, 
COVER offers valuable insights into how structured reasoning can enhance the robustness of video understanding by constructing a sub-question–based counterfactual video QA benchmark across multiple levels of abstraction and thoroughly evaluating mainstream MLLMs’ logical reasoning abilities.

\definecolor{darkgreen}{rgb}{0.0, 0.5, 0.0}
\definecolor{lightgray}{rgb}{0.9, 0.9, 0.9}

\begin{table*}[t]
    \centering
    \renewcommand{\arraystretch}{1.1}
    \setlength{\tabcolsep}{10pt}
    \scalebox{0.8}{
    \begin{tabular}{l|ccccccc}
    \toprule 
     \textbf{Benchmark} & \textbf{Video} & \textbf{Q\&A} & \textbf{Qs Source} & \textbf{CF} & \textbf{SQP} & \textbf{PCD} & \textbf{ACD}  \\
     \midrule
     CoFCA~\citep{cofca} & \color{red}\ding{55} & \color{darkgreen}\Checkmark  & H\&A & \color{darkgreen}\Checkmark & \color{darkgreen}\Checkmark & \color{red}\ding{55} & \color{red}\ding{55} \\ 
     CFMM~\citep{cfmm} & \color{red}\ding{55} & \color{darkgreen}\Checkmark  & H & \color{darkgreen}\Checkmark  & \color{red}\ding{55} & \color{red}\ding{55} & \color{red}\ding{55} \\ 
     Video-MME~\citep{videomme} & \color{darkgreen}\Checkmark & \color{darkgreen}\Checkmark  & H & \color{red}\ding{55} & \color{red}\ding{55} & \color{darkgreen}\Checkmark & \color{red}\ding{55} \\ 
     CRIPP-VQA~\citep{cripp} & \color{darkgreen}\Checkmark & \color{darkgreen}\Checkmark  & H & \color{darkgreen}\Checkmark & \color{red}\ding{55}  & \color{red}\ding{55} & \color{red}\ding{55} \\ 
     VITATECS~\citep{vitatecs} & \color{darkgreen}\Checkmark & \color{red}\ding{55}  & H\&A & \color{darkgreen}\Checkmark & \color{red}\ding{55} & \color{red}\ding{55} & \color{red}\ding{55} \\ 
     \midrule
     COVER~(ours) & \color{darkgreen}\Checkmark & \color{darkgreen}\Checkmark  & H\&A & \color{darkgreen}\Checkmark  & \color{darkgreen}\Checkmark & \color{darkgreen}\Checkmark & \color{darkgreen}\Checkmark \\ 
    \bottomrule 
    \end{tabular}}
    \caption{Comparison with existing benchmarks. \textbf{Video}: whether the benchmark involves video data; \textbf{Q\&A}: whether it follows a question-and-answer format; \textbf{Qs source}: H indicates human annotation, A indicates automatic annotation; \textbf{CF}: whether counterfactual questions are included; \textbf{PCD}: whether the benchmark is categorized by the model's perceptual and cognitive demands; \textbf{ACD}: whether tasks are divided based on object abstraction (abstract vs. concrete).}
    \label{tab:comparsion}
\end{table*}

\section{Related Work}

\noindent \textbf{Multimodal Large Language Models and Their Evaluation.}  
Recent advances in MLLMs have greatly improved their capacity to understand and reason over diverse modalities, such as images, text, and videos. To evaluate these models, benchmarks targeting static image comprehension have emerged, including MM-Vet~\cite{mm-vet}, MME~\cite{mme}, MMBench~\cite{mmbench}, and GQA~\cite{gqa}. These primarily assess visual recognition and spatial reasoning. Extending beyond static content, video-centric benchmarks like Video-MME~\cite{videomme}, MvBench~\cite{Mvbench}, and SEED-Bench~\cite{seed-bench} focus on temporal dynamics and contextual reasoning. Together, these benchmarks reflect the growing demand for evaluating multimodal understanding in both static and dynamic environments.

\noindent \textbf{Chain-of-Thought and Counterfactual Reasoning in MLLMs.}  
Chain-of-Thought (CoT) reasoning~\cite{CoT} enhances logical inference by breaking down complex tasks into intermediate steps. Multimodal adaptations~\cite{m-CoT, ddcot} extend this strategy across modalities, showing gains in structured reasoning. Counterfactual reasoning, which examines hypothetical changes and their consequences, has also gained traction. Prior work explores this in text~\cite{ror,cofca}, visual QA~\cite{cfmm}, and hybrid settings. ACQUIRED~\cite{acquired} proposes a taxonomy of counterfactual types, while AuroraCap~\cite{AuroraCap} and CoFCA~\cite{cofca} assess models' sub-task decomposition and multi-step reasoning. These approaches collectively underscore the importance of structured, causal reasoning in complex multimodal tasks.

\noindent \textbf{Multimodal Generalization and Video Counterfactual Benchmarks.}  
Although several benchmarks target video-based counterfactual understanding—such as CRIPP-VQA for physical properties, VITATECS for captioning, and ACQUIRED for scenario taxonomy~\cite{vitatecs,cripp}—they remain narrow in scope. Most fail to capture the breadth of reasoning demands in real-world counterfactual scenarios.

\noindent To address this, \textsc{COVER} introduces a fine-grained framework for evaluating counterfactual video reasoning via sub-question decomposition. It explicitly distinguishes between abstract vs. concrete object attributes and perceptual vs. cognitive reasoning demands. As summarized in Table~\ref{tab:comparsion}, \textsc{COVER} broadens the evaluation spectrum, enabling a more nuanced and comprehensive assessment of multimodal counterfactual reasoning than prior efforts.

\section{The COVER Benchmark}

\begin{figure}[ht]
\centering
  \includegraphics[width=0.45\textwidth]{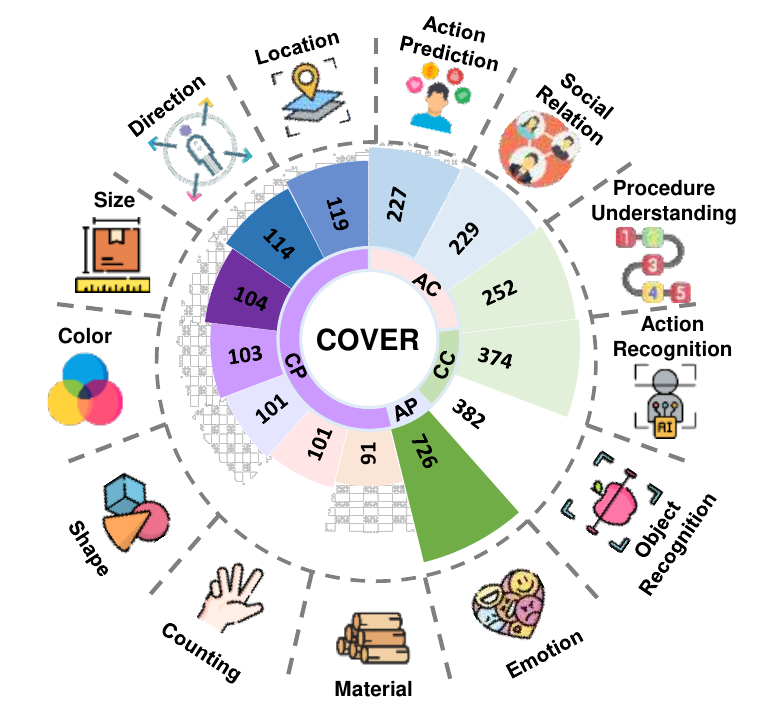}
  \caption{Overview of the 13 tasks in COVER. Numbers on the outer edge of the rose chart indicate the total number of question pairs for each task, while inner labels denote the corresponding dimension: \textbf{A\&C} (Abstract Cognition), \textbf{C\&C} (Concrete Cognition), \textbf{A\&P} (Abstract Perception), and \textbf{C\&P} (Concrete Perception).}
  \label{fig:category}
\end{figure}

This section provides a comprehensive overview of the construction of \textbf{COVER}. We introduce our data partitioning framework designed to evaluate MLLM reasoning ability across four complementary dimensions. Next, we describe the data curation process, which domain experts have rigorously validated to ensure the high quality and reliability of the benchmark.

Our benchmark includes approximately 2,800 videos, which are paired with around 12,000 to 13,000 individual QA instances. As stated in L-Figure \ref{fig:enlarge}, the enhanced version of our dataset consists of about 2.9k question pairs, with each pair comprising at least three individual QA items: one original question, one counterfactual question, and at least one sub-question (often multiple). 

\subsection{Benchmark Definition}

As illustrated in Figure~\ref{fig:category}, we categorize the 13 benchmark tasks into four quadrants based on the abstract-concrete and perceptual-cognitive dimensions.
\textbf{Abstract-Perception:} (1) Emotion: Understanding and recognizing emotional states.
\textbf{Concrete-Perception:} (2) Counting: Quantity recognition and calculation. (3) Color: Perceiving object colors. (4) Direction: Sensing motion trends. (5) Size: Identifying object dimensions. (6) Shape: Perceiving object shapes. (7) Material: Recognizing object materials. (8) Location: Detecting object positions.
\textbf{Concrete-Cognition:} (9) Action Recognition: Identifying specific actions. (10) Object Recognition: Recognizing specific objects.
\textbf{Abstract-Cognition:} (11) Action Prediction: Forecasting future actions. (12) Procedure Understanding: Comprehending sequential processes and logic. (13) Social Relation: Understanding social relationships.

\noindent \textbf{Division of Abstract and Concrete Scenes.}
This distinction reflects a functional differentiation within cognitive representation systems. Neuroscientific studies~\cite{abstract_concrete} suggest that concrete concepts rely heavily on multi-modal perceptual simulations (e.g., object shape, material), while abstract concepts are primarily represented through language-mediated symbolic operations. Abstract tasks often require integrating non-perceptual information, such as contextual encoding for emotion recognition or constructing temporal causal models for action prediction. 

\noindent \textbf{Division of Perception and Cognition.}
Perception involves the initial reception of external stimuli through sensory organs, converting them into neural signals that provide raw environmental data. Cognition, built upon perception, refers to the further processing of these signals, encompassing higher-level mental functions such as memory, attention, language comprehension, problem-solving, and reasoning. This distinction underscores different stages of information processing, with perception forming the foundation upon which cognitive functions are built.

\subsection{Data Construction}

\begin{figure}[ht]
\centering
  \includegraphics[width=0.48\textwidth]{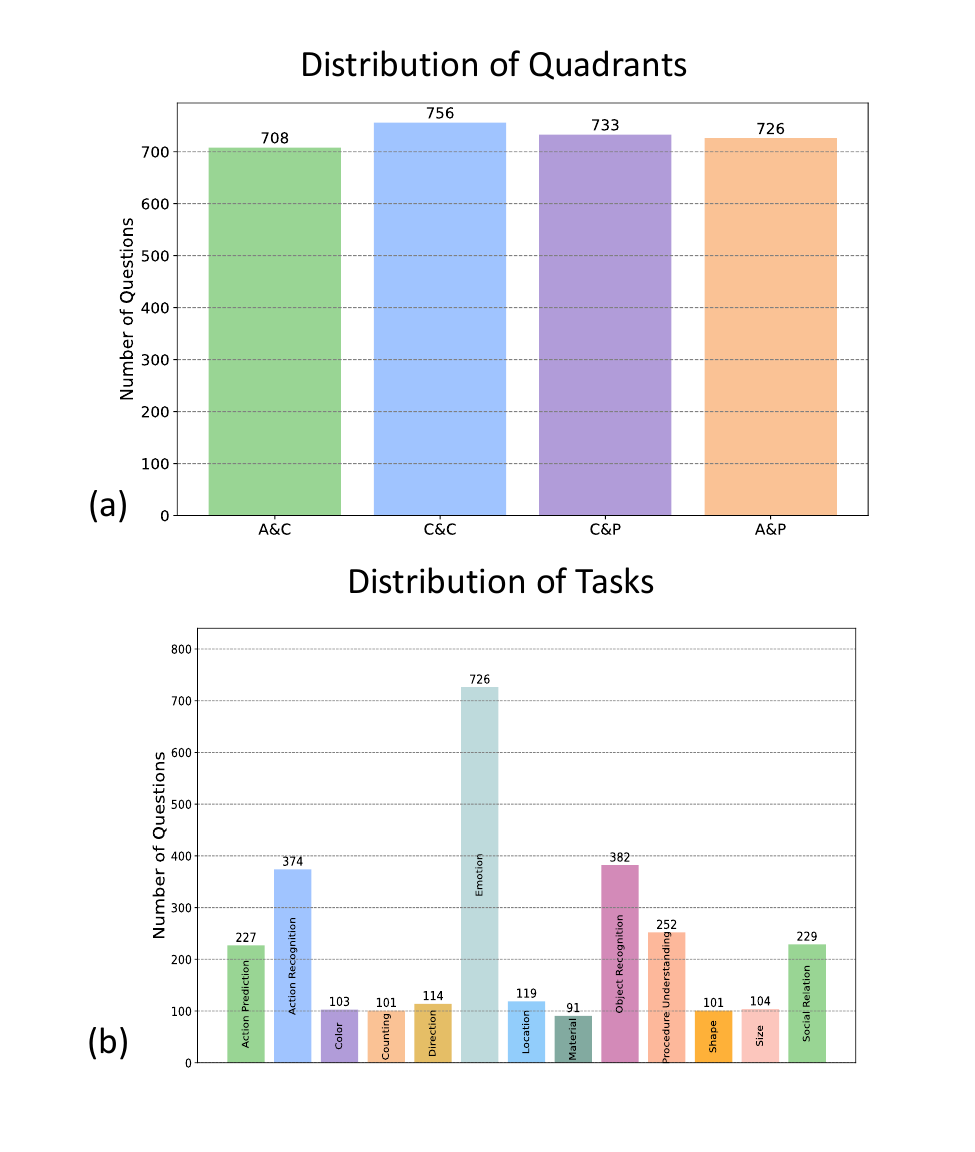}
  \caption{(a) Distribution of question pairs across the four quadrants. (b) Distribution of question pairs across the 13 tasks.}
  \label{fig:num_staticis}
\end{figure}

To construct COVER, we carefully selected a diverse range of open-source and research-available video sources, including ~\citet{Charades_v1_480, clevrer, FunQA, MSRVTTandMSVD_Zero_Shot_QA, nturgbd, Perception_test, MoVQA(scene_qa), sta, TGIFQA, VATEX, vlnqa}. These sources encompass various real-world scenarios, ranging from daily activity recognition to complex scene understanding. As shown in Appendix Figure \ref{fig:enlarge}, we collected 146 videos and designed 150 aspect-specific QA pairs, each of which underwent dual-team review for validation. To ensure balanced coverage across the four quadrants, we expanded the seed data using GPT-generated instances (720-760 per quadrant) to mitigate any potential biases. The detailed statistical findings are comprehensively presented in Figure \ref{fig:num_staticis}. The frame count of videos in \textbf{COVER} ranges from 16 to 1739, with an average of 294.34 frames. 
We finally constructed 2,923 high-quality counterfactual question-answer pairs. Each question-answer pair consists of an original question, which presents no hypothetical context, and a counterfactual question, which introduces situational assumptions and sub-questions that enable granular reasoning analysis. 

Eight annotators further validated the dataset and checked logical consistency to ensure the reasoning relied solely on the video content. Additionally, three experts cross-validate the dataset (see Appendix Table \ref{tab:cross_val}) to confirm the structural balance.

    

\begin{table}
    \centering
    \small
    \begin{tabular}{lccc}
    \toprule
     & \makecell{\textbf{$ori_{acc}$}} & \makecell{\textbf{$cf_{acc}$}} & \makecell{\textbf{$sub_{acc}$}} \\
    \midrule
    GPT-4o  &70.26 & 45.93 & 56.94 \\
    GPT-4o-mini &67.32 & 51.47 & 55.94 \\
    Claude-3.5-Sonnet &63.60 & 38.04 & 49.40 \\
    Gemini-1.5-Pro &74.82 & 49.64 & \underline{63.76} \\
    Gemini-1.5-Flash &73.90 & 48.75 & 62.52 \\
    Gemini-2.0-Flash &\textbf{77.18} & 46.90 & 62.92 \\
    \midrule
    InternVL2.5-78B &\underline{76.74} & \textbf{59.46} & \textbf{67.23} \\
    LlaVA-Video-72B  &64.35 & 56.04 & 61.54 \\
    InternVL2.5-26B &75.40 & 51.08 & 62.65 \\
    InternVL2.5-8B &74.31 & \underline{57.75} & 61.65 \\
    VideoLlama3-8B &73.04 & 51.25 & 60.09 \\
    LlaVa-OV-7B &62.74 & 51.80 & 56.42 \\
    LLaVA-Video-7B &60.52 & 51.93 & 55.11 \\
    Qwen2-VL-7B &71.83 & 46.90 & 58.40 \\
    VILA-U-7B &60.01 & 38.42 & 47.32 \\
    VILA1.5-7B &60.25 & 57.34 & 53.18 \\
    \bottomrule
    \end{tabular}
    \caption{General assessment results of COVER. $ori_{acc}$, $cf_{acc}$, and $sub_{acc}$ denote the accuracies of the original, counterfactual, and sub-questions, respectively.}

    \label{tab:eval_results}
    \vspace{-2mm}
\end{table}

\section{Experiments}

In this section, we systematically evaluate MLLMs of varying scales on the COVER benchmark to foster transparent and reproducible research. Our evaluation spans four key dimensions: cognition, perception, abstraction, and concreteness. It encompasses diverse reasoning sub-tasks, including counterfactual reasoning, direct inference, and sub-question-guided reasoning. We compare both open-source and proprietary models across different parameter scales to analyze their relative strengths and limitations. We begin by detailing the experimental setup.

\subsection{Settings}

To ensure a thorough evaluation, we selected a diverse set of representative MLLMs, including commercial closed-source models such as GPT-4o~\cite{gpt4o}, Claude~\cite{claude35}, and Gemini~\cite{gemini15}, as well as leading open-source models such as InternVL2.5~\cite{internvl2.5}, LLAVA-Video~\cite{llava-video}, LLaVA-OV~\cite{llavaonevision}, Qwen2-VL~\cite{qwen2-vl}, VideoLLaMA3~\cite{videollama3}, vila-u~\cite{vila-u}, and VILA~\cite{vila}. These models span a wide range of parameter scales and design paradigms, offering a comprehensive view of the current landscape in multimodal learning.

We evaluate model performance on video understanding using three metrics: $ori_{acc}$ (original question accuracy), $cf_{acc}$ (counterfactual question accuracy), and $sub_{acc}$ (sub-question accuracy), with scores averaged over at least three runs. All models are tested under identical conditions, using a consistent frame extraction strategy that samples 16 frames per video segment. The impact of alternative sampling strategies is discussed in Chapter~5.

\begin{table*}
\centering
\begin{adjustbox}{max width=\linewidth}
\begin{tabular}{l!{\vrule width \lightrulewidth}cccccccccccc} 
\toprule
\multicolumn{1}{c|}{\multirow{2}{*}{\textbf{Models}}} & \multicolumn{3}{c}{\textbf{A\&C (\%)}} & \multicolumn{3}{c}{\textbf{C\&C (\%)}} & \multicolumn{3}{c}{\textbf{C\&P (\%)}} & \multicolumn{3}{c}{\textbf{A\&P (\%)}}  \\ 
\cmidrule(lr){2-4}\cmidrule(lr){5-7}\cmidrule(lr){8-10}\cmidrule(lr){11-13}
& $ori_{acc}$      & $cf_{acc}$   & $sub_{acc}$              & $ori_{acc}$      & $cf_{acc}$   & $sub_{acc}$      & $ori_{acc}$      & $cf_{acc}$   & $sub_{acc}$   & $ori_{acc}$      & $cf_{acc}$   & $sub_{acc}$       \\ 

\midrule

GPT-4o & 71.05 & 41.81 & 41.70 & 74.87 & 43.65 & 68.36 & 69.95 & 42.62 & 50.52 & 65.01 & 55.65 & 63.97 \\
GPT-4o-mini & 62.29 & 52.40 & 42.97 & 76.32 & 54.37 & 66.49 & 64.62 & 40.85 & 44.78 & 65.56 & \textbf{58.26} & 65.96 \\
Claude-3.5-sonnet & 56.92 & 37.01 & 35.55 & 70.11 & 42.33 & 61.77 & 60.03 & 32.88 & 40.08 & 66.94 & 39.81 & 56.81\\
Gemini 1.5 Pro & 69.49 & 44.49 & \underline{53.36} & \underline{82.14} & 51.98 & 72.78 & 71.76 & 43.93 & 56.81 & 75.48 & 57.99 & 69.54\\
Gemini 1.5 Flash & 70.48 & 45.34 & 52.23 & 82.01 & 49.34 & 71.51 & 70.67 & 42.02 & 51.90 & 72.04 & \textbf{58.26} & \textbf{71.36}\\
Gemini 2.0 Flash & \textbf{74.29} & 44.36 & 51.38 & \textbf{83.99} & 47.75 & \underline{72.84} & 74.22 & 38.74 & 58.26 & \underline{75.90} & 57.71 & 66.84\\

\midrule
InternVL2.5-78B & \underline{72.88} & \textbf{59.60} & \textbf{57.67} & 80.95 & \underline{63.62} & \textbf{75.62} & \underline{75.99} & \underline{58.25} & \textbf{63.65} & \textbf{76.86} & 56.20 & \underline{70.07} \\
LLaVA-Video-72B & 53.11 & 54.94 & 53.14 & 65.34 & 60.45 & 67.03 & 67.94 & 52.39 & 53.49 & 70.66 & 56.20 & 70.01 \\
InternVL2.5-26B & 71.05 & 47.74 & 50.53 & 80.95 & 58.99 & 72.17 & \textbf{76.13} & 47.20 & \underline{60.12} & 73.14 & 50.00 & 65.61 \\
InternVL2.5-8B & 69.77 & \underline{58.62} & 49.96 & 80.95 & \textbf{64.55} & 71.02 & 73.94 & 55.80 & 54.66 & 72.18 & 51.79 & 68.19 \\
VideoLLama3-8B & 68.08 & 45.62 & 49.68 & 81.35 & 54.89 & 68.36 & 72.99 & 50.75 & 51.62 & 69.28 & 53.44 & 67.90 \\
LLaVA-ov-7B & 54.66 & 51.69 & 47.49 & 62.96 & 53.04 & 61.77 & 64.53 & 49.66 & 49.48 & 68.60 & 52.75 & 64.73 \\
LLaVA-Video-7B & 50.14 & 55.23 & 44.52 & 61.64 & 50.53 & 60.50 & 63.57 & 52.52 & 49.97 & 66.39 & 49.59 & 63.03 \\
Qwen2-VL-7B & 65.96 & 49.15 & 48.41 & 82.14 & 43.39 & 67.03 & 71.21 & 45.57 & 50.52 & 67.49 & 49.72 & 65.02 \\
VILA-U-7B & 58.19 & 39.83 & 38.87 & 63.10 & 41.93 & 54.51 & 59.07 & 37.93 & 37.94 & 59.50 & 33.88 & 55.34 \\
VILA1.5-7B & 54.80 & 55.93 & 39.29 & 66.93 & 62.30 & 63.52 & 55.25 & \textbf{58.53} & 44.64 & 63.64 & 52.34 & 61.91 \\

\bottomrule
\end{tabular}
\end{adjustbox}
\caption{Performance of MLLMs on COVER, based on our quadrant formulation (A\&C, C\&C, C\&P, A\&P), measured by original, counterfactual, and sub-question accuracy.}
\label{tab:eval-quart}
\end{table*}

\subsection{Main Results}

As shown in Table \ref{tab:eval_results}, Gemini-2.0-Flash ($ori_{acc}$ 77.18\%) and InternVL2.5-78B ($ori_{acc}$ 76.74\%) rank as the top two models, demonstrating their strong foundational video understanding. The lower scores of VILA-U-7B ($ori_{acc}$ 60.01\%) and LLaVA-Video-7B ($ori_{acc}$ 60.52\%) highlight the limitations of smaller models in processing long sequences.
InternVL2.5-78B ($cf_{acc}$ 59.46\%) shows significant dominance in handling conditional reasoning and complex contexts. Notably, counterfactual questions cause sharp accuracy drops compared to the original questions: GPT-4o (-24.33\%) and Gemini-1.5-Pro (-25.18\%), indicating that most models struggle with counterfactual reasoning. 

Most models exhibit higher $sub_{acc}$ than $cf_{acc}$ (e.g., Claude-3.5-Sonnet 49.40\% vs. 38.04\%, LLaVA-Video-72B 61.54\% vs. 56.04\%). This suggests better stability in localized reasoning tasks than in holistic tasks, where error accumulation impacts performance. In the Appendix, we provide detailed case demonstrations in Figure \ref{fig:case1}.

\textbf{Closed-source Model Performance.} As shown in Table \ref{tab:eval-quart}, Gemini 1.5 Pro demonstrates strong dominance in both concrete cognition ($ori_{acc}$ 82.14\%) and abstract perception tasks ($ori_{acc}$ 75.48\%). Gemini 2.0 Flash excels in abstract perception ($ori_{acc}$ 75.90\%) and concrete perception tasks ($ori_{acc}$ 74.22\%), showcasing strong capabilities in handling high-complexity perceptual tasks.

\textbf{Open-source Model Performance.} As shown in Table~\ref{tab:eval-quart}, InternVL2.5-78B leads in abstract cognition ($ori_{acc}$ 72.88\%) and concrete perception tasks ($cf_{acc}$ 58.25\%), reflecting a deep understanding of abstract concepts and complex logic. Lightweight models like Qwen2-VL-7B perform well in concrete cognition ($ori_{acc}$ 82.14\%) but face limitations in abstract tasks ($ori_{acc}$ 65.96\% in A\&C) due to their smaller parameter size, revealing distinct capabilities across model types.
Commercial models, such as the Gemini series, maintain strong performance in concrete cognition and abstract perception tasks but generally fall behind open-source models in counterfactual reasoning. Most models struggle with counterfactual reasoning, with only InternVL2.5-7BB and VILA1.5-7B showing some task-specific advantages, highlighting the need for targeted optimization in conditional hypothesis modeling.

\subsection{Sub-question Guideline}

\begin{table*}
\small
\centering
\begin{tabular}{l|c|c|cc}
    \Xhline{1.0pt}
    \multirow{2}{*}{\textbf{Model}} & \multicolumn{1}{c|}{\textbf{Without CoT}} & \multicolumn{1}{c|}{\textbf{With CoT}} & \multicolumn{2}{c}{\textbf{Guide-CoT}} \\
    ~ & \textbf{$cf_{acc}$} & \textbf{$cf_{acc}$} &  \textbf{$cf_{acc}$} & \textbf{$cf_{with ans}$} \\
    \Xhline{0.8pt}
    GPT-4o-mini & 51.47 & \underline{58.62} &  \underline{57.93} & \underline{68.07}  \\
    InternVL2.5-78B & \textbf{59.46} & \textbf{60.42} &  \textbf{58.33} & \textbf{68.29}  \\
    LlaVA-Video-72B  & 56.04 &56.24 &  53.51 & 63.12  \\
    InternVL2.5-8B & \underline{57.75} &57.06 &  52.41 & 57.75  \\
    VideoLlama3-8B & 51.25 &52.82 &  53.06 & 52.79  \\
    LLaVA-Video-7B & 51.93 &51.42 &  51.39 & 54.12  \\
    Qwen2-VL-7B & 46.90 &50.36 &  45.71 & 50.88  \\
    \Xhline{1.0pt}
    
\end{tabular}
\caption{Comparison between CoT and Guide-CoT performance across MLLMs on the COVER benchmark.}
\label{tab:cot-comparsion}
\end{table*}

We propose Guide-CoT to study the influence of different reasoning paths on model performance through human-annotated sub-problems. We design comparative experiments between CoT and Guide-CoT to analyze how automatically generated sub-questions from CoT versus manually annotated sub-questions affect model reasoning capabilities.

Comparing the Without CoT and CoT approaches based on Table~\ref{tab:cot-comparsion}, we find that the $cf_{acc}$ of most models under CoT significantly exceeds the Without CoT baseline, such as Qwen2-VL-7B (+3.46\%) and GPT-4o-mini (+7.15\%), which indicates that CoT enhances reasoning processes, particularly in more complex tasks.

However, examining Guide-CoT results reveals that manually designed sub-questions may not always lead to substantial improvement over automatically generated ones, as seen with GPT-4o-mini's $cf_{acc}$ of 57.93\% under Guide-CoT, slightly lower than the 58.62\% under CoT. This does not imply the ineffectiveness of manual sub-questions but suggests that model behaviors may not always align with human-designed reasoning paths, potentially due to task complexity or the nature of the sub-questions themselves. We hypothesize that manually provided sub-questions could introduce extraneous patterns or "pseudo-features" that are not directly relevant to the reasoning task, leading to a subtle reduction in performance.

The $cf_{withans}$ column in Guide-CoT indicates sub-questions that include standard answers. For InternVL2.5-78B, $cf_{withans}$ reaches 68.29\%, reflecting an 8.63\% improvement over the no-CoT baseline, in contrast to CoT’s modest gain of only 0.96\% (from 59.46\% to 60.42\%). This suggests that providing complete answers substantially enhances reasoning accuracy, particularly in complex or multi-step tasks. Standard-answer sub-questions enable the model to better integrate information and verify intermediate reasoning steps, resulting in improved consistency and overall performance. Detailed case studies are presented in Appendix Figure~\ref{fig:case2} to further illustrate these findings and analyze the interplay between reasoning paths and task complexity.

The results from our experiments strongly support the notion that reasoning plays a pivotal role in model robustness and generalization. Our study extends these insights by demonstrating that multimodal models, especially in the context of video tasks, rely heavily on robust reasoning capabilities for effective generalization. The significant performance improvements observed with counterfactual reasoning and sub-question decomposition highlight that models’ ability to handle complex, conditional, and dynamic contexts is crucial for their robustness, a finding not fully explored in prior research.

\begin{table*}[!ht]
\small
\centering
\begin{adjustbox}{max width=\linewidth}
\begin{tabular}{c!{\vrule width \lightrulewidth}cccccccccccccccccc} 
\toprule
\multicolumn{1}{c|}{\multirow{2}{*}{\textbf{Frames}}} & \multicolumn{3}{c}{\textbf{InternVL2.5-1B}} & \multicolumn{3}{c}{\textbf{InternVL2.5-2B}} & \multicolumn{3}{c}{\textbf{InternVL2.5-4B}} & \multicolumn{3}{c}{\textbf{InternVL2.5-8B}} \\ 
\cmidrule(lr){2-4}\cmidrule(lr){5-7}\cmidrule(lr){8-10}\cmidrule(lr){11-13}
& $ori_{acc}$ & $cf_{acc}$ & $sub_{acc}$ & $ori_{acc}$ & $cf_{acc}$ & $sub_{acc}$     & $ori_{acc}$ & $cf_{acc}$ & $sub_{acc}$ & $ori_{acc}$ & $cf_{acc}$ & $sub_{acc}$   \\ 

\midrule
2  & 66.16 & 35.61 & 55.27 & 65.31 & 44.20 & 54.99 & 72.56 & 48.31 & 60.88 & 71.26 & 58.50 & 60.07 \\
4  & 68.32 & 34.72 & 55.52 & 68.83 & 42.11 & 58.84 & 74.41 & 46.49 & 61.79 & 73.35 & 58.47 & 60.96 \\
8  & 68.94 & 35.10 & 55.11 & 68.22 & 41.43 & 55.75 & 75.03 & 45.60 & 61.79 & 74.14 & 57.06 & 61.60 \\
16  & 69.76 & 35.89 & 55.19 & 70.07 & 40.68 & 55.49 & 75.61 & 45.23 & 61.63 & 74.31 & 57.75 & 61.65 \\
32  & 69.04 & 36.50 & 55.04 & 70.13 & 39.69 & 55.48 & 75.54 & 45.09 & 60.96 & 74.10 & 57.03 & 61.42 \\
64  & 68.18 & 37.39 & 54.80 & 68.90 & 40.06 & 55.44 & 74.41 & 46.56 & 60.70 & 74.20 & 58.09 & 61.30 \\
\bottomrule
\end{tabular}
\end{adjustbox}
\caption{Performance of MLLMs on COVER using different frame sampling strategies. The frame selection follows standard practices in video QA benchmarks, where the number of input frames is set to $\min(\text{video length}, \text{predefined sampling count})$.}
\label{tab:frame}
\end{table*}

\section{Analysis}
In this chapter, we begin by analyzing the impact of video frame sampling rates on MLLMs' video understanding and reasoning abilities. We then proceed with an in-depth examination of MLLMs' robustness and logical reasoning performance.


\subsection{Ablation Study of Video Frames}

As shown in Table~\ref{tab:frame}, as the parameter size of LLMs increases, there is a rising trend in $ori_{acc}$, $cf_{acc}$, and $sub_{acc}$. For instance, with 16 frames, the InternVL2.5-1B model achieves $ori_{acc}$, $cf_{acc}$, and $sub_{acc}$ of 69.76\%, 35.89\%, and 55.19\% respectively. The InternVL2.5-2B scores 70.07\%, 40.62\%, and 55.49\%, while the InternVL2.5-4B reaches 75.61\%, 45.23\%, and 61.68\%, indicating that larger LLMs have enhanced capabilities in handling complex problems.
Under the same vision tower settings, $ori_{acc}$ shows a clear upward trend as the number of frames increases. For example, the InternVL2.5-8B's $ori_{acc}$ rises from 71.26\% at 2 frames to 74.20\% at 64 frames. However, $cf_{acc}$ tends to decrease with more frames. The InternVL2.5-2B's $cf_{acc}$ drops from 44.20\% at 2 frames to 40.06\% at 64 frames.
Models with more parameters generally perform better in $ori_{acc}$, $cf_{acc}$, and $sub_{acc}$, highlighting the significant role of LLMs in multimodal reasoning. Additionally, increasing visual information (by raising the frame count) can enhance $ori_{acc}$, but excessive visual information, especially in complex or counterfactual reasoning scenarios, may impair the model's reasoning ability, leading to a decline in $cf_{acc}$.

\subsection{Robustness and Logical Reasoning in MLLMs}

\begin{figure*}[!ht]
\centering
  \includegraphics[width=1.0\textwidth]{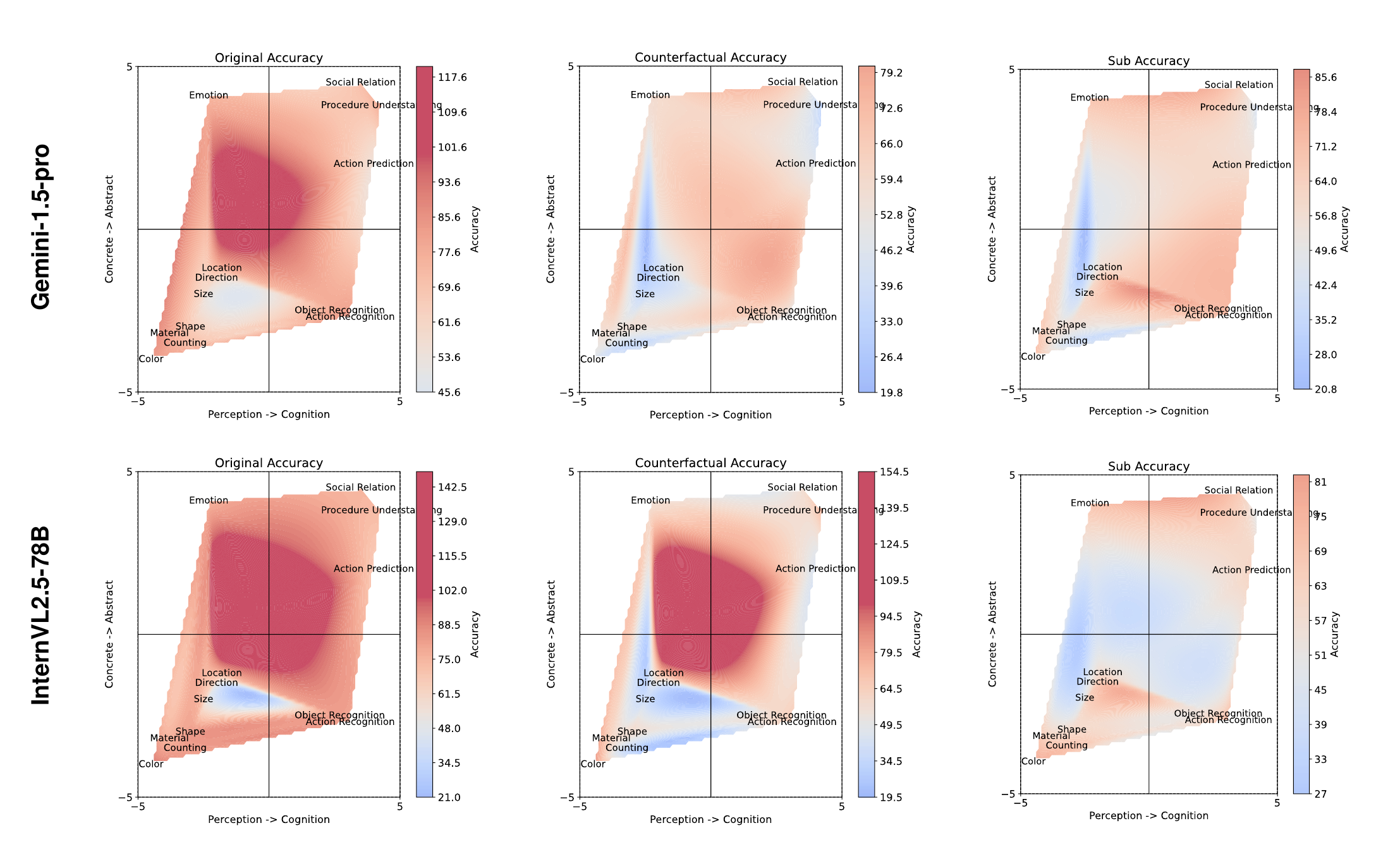}
  \caption{Heatmaps of task performance for Gemini-1.5-pro and InternVL2.5-78B, using hollow circles to depict task distributions across the four quadrants. The top three panels show results for Gemini-1.5-pro, and the bottom three for InternVL2.5-78B. \textbf{Left}: Accuracy on original questions. \textbf{Middle}: Performance on counterfactual questions. \textbf{Right}: Accuracy on sub-questions. A gradient color bar—from azure (low accuracy) to crimson (high accuracy)—is placed along the right margin of each heatmap to indicate performance levels.}
  \label{fig:internvl_example}
\end{figure*}

\begin{figure*}[h!]
\centering
  \includegraphics[width=0.8\textwidth]{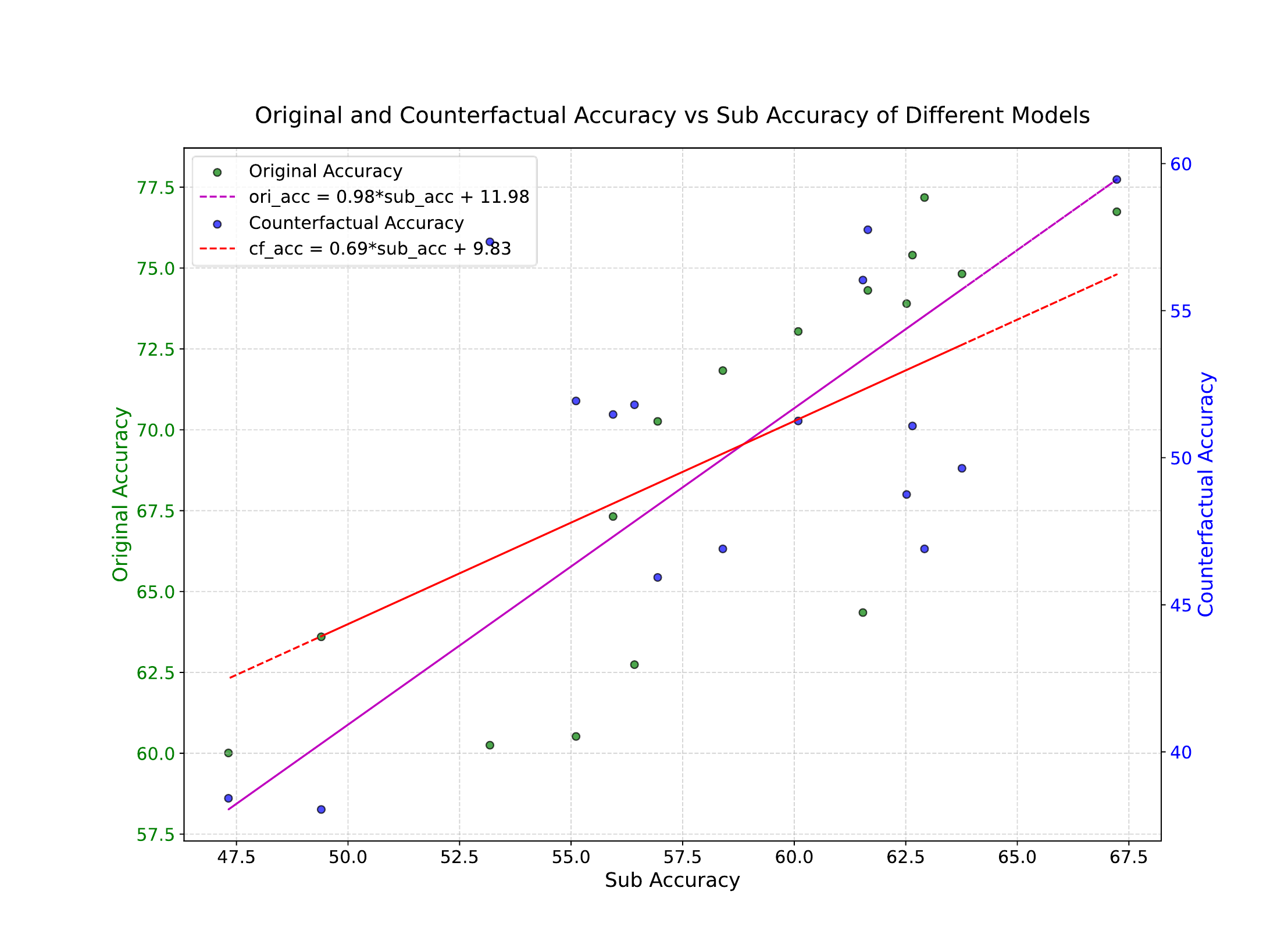}
  \caption{Scatter plot showing correlations among $ori_{acc}$, $sub_{acc}$, and $cf_{acc}$ across models. The red line represents the linear function fitted between $ori_{acc}$ and $sub_{acc}$, while the purple line represents the linear function fitted between $cf_{acc}$ and $sub_{acc}$.}
  \label{fig:acclinechart}
\end{figure*}

The ability of MLLMs to answer original questions serves as a key indicator of their overall understanding capabilities, while performance on sub-questions reveals single-step reasoning proficiency.
A notable observation is the strong Pearson correlation between $ori_{acc}$ and $sub_{acc}$ reaches 0.836, indicating a strong connection between model understanding and reasoning capabilities. Furthermore, as shown in Figure \ref{fig:acclinechart}, the correlation between $sub_{acc}$ and $cf_{acc}$ is 0.608.
These moderately strong correlations indicate that a model's ability to comprehend original questions plays a fundamental role in enabling effective step-by-step reasoning. Similarly, the correlation between \( ori_{acc} \) and \( sub_{acc} \) suggests that models with a higher understanding capability tend to perform better when solving decomposed sub-questions, reinforcing the notion that comprehension and reasoning are interdependent. However, the moderate correlation between \( sub_{acc} \) and \( cf_{acc} \) suggests that counterfactual reasoning involves additional complexities, making it a more challenging task than single-step reasoning.

As illustrated in Table \ref{tab:model_performance_gap}, We observed that across multiple models, the probability $P(\text{cf\_right} | \text{sub\_right})$ was significantly higher than $P(\text{cf\_right} | \text{sub\_wrong})$, clearly indicating that the correctness of sub-questions is a strong predictor of overall counterfactual performance. 

\begin{table*}[htbp]
\centering
\small
\begin{tabular}{lcccc}
\toprule
Model & P($cf_{right}$|$sub_{right}$) & P($cf_{wrong}$|$sub_{right}$) & P($cf_{right}$|$sub_{wrong}$) & P($cf_{wrong}$|$sub_{wrong}$) \\
\midrule
gemini-1.5-pro & 56.54 & \textbf{43.45} & 44.99 & \textbf{55.01} \\
GPT-4o-mini & 59.49 & 40.51 & 47.65 & 52.35 \\
InternVL2.5-78B & 62.90 & 37.10 & \textbf{56.67} & 43.34 \\
LlaVA-Video-72B & \textbf{63.28} & 36.72 & 51.60 & 48.40 \\
\bottomrule
\end{tabular}
\caption{Conditional probabilities of counterfactual accuracy given sub-question outcomes. P($cf_{right}$\,|\, $sub_{right}$) and P($cf_{wrong}$\,|\, $sub_{right}$) denote the likelihood of answering the counterfactual question correctly or incorrectly when the sub-question is correct; similarly, P($cf_{right}$\,|\, $sub_{wrong}$) and P($cf_{wrong}$\,|\, $sub_{wrong}$) apply when the sub-question is incorrect.}

\label{tab:model_performance_gap}
\end{table*}

Analysis of the heat maps in Figure~\ref{fig:internvl_example} reveals different performance patterns in the quadrants, highlighting the interaction between comprehension, step-by-step reasoning, and counterfactual inference. In abstract reasoning tasks such as social inference and procedural understanding, the drop from \( sub_{acc} \) to \( ori_{acc} \) is minimal, and the transition to \( cf_{acc} \) remains stable. This suggests that models can effectively leverage sub-question reasoning and maintain performance even under counterfactual assumptions. In contrast, the \textbf{concrete perception quadrant}—involving tasks like object recognition and motion understanding—shows a sharper decline from \( sub_{acc} \) to \( ori_{acc} \), and further to \( cf_{acc} \). This indicates that perception-heavy tasks pose greater challenges, as models struggle to decompose complex sensory input into reasoning steps required for counterfactual understanding.

Overall, our findings indicate that counterfactual reasoning is inherently more challenging than single-step reasoning, especially in perception-intensive tasks where models must infer causality beyond pattern recognition. In contrast, the relatively stable gap between \( sub_{acc} \) and \( cf_{acc} \) in abstract-cognitive tasks suggests that models can better leverage conceptual knowledge. Enhancing counterfactual reasoning in perception-heavy scenarios remains a key challenge, likely requiring improved causal inference and reasoning mechanisms.

\subsection{The Effects of Model Scale}
We conduct systematic analyses to characterize performance gaps across original, counterfactual, and sub-question accuracies. Our goal is to mitigate these gaps by examining factors such as model scale, training alignment, and reasoning strategies. As shown in Table~\ref{tab:three_acc_gap}, with similar visual backbones, increasing language model size significantly reduces the performance gap—particularly between sub-question and counterfactual accuracy. Specifically, the absolute difference between $\text{ori}_{\text{acc}}$ (70.07\%) and $\text{cf}_{\text{acc}}$ (40.68\%) is 29.39\% for the 2B model, increases slightly to 30.38\% for the 4B model, and then drops substantially to 16.56\% for the 8B model. Similarly, the gap between $\text{cf}_{\text{acc}}$ and $\text{sub}_{\text{acc}}$ grows from 14.81\% (2B) to 16.40\% (4B), before narrowing sharply to 3.90\% (8B).

\begin{table}[htbp]
\centering
\small
\begin{tabular}{lccc}
\toprule
Model & $ori_{acc}$ & $cf_{acc}$ & $sub_{acc}$ \\
\midrule
InternVL2.5-2B & 70.07 & 40.68 & 55.49 \\
InternVL2.5-4B & 75.61 & 45.23 & 61.63 \\
InternVL2.5-8B & 74.31 & 57.75 & 61.65 \\
\bottomrule
\end{tabular}
\caption{Variations in three accuracy metrics across different model sizes.}
\label{tab:three_acc_gap}
\end{table}

\section{Conclusion}
We introduce COVER, a comprehensive benchmark for counterfactual video reasoning that evaluates MLLMs across four dimensions: abstract-concrete and perception-cognition. By decomposing complex queries into structured sub-questions, COVER enables fine-grained analysis and reveals a strong correlation between sub-question accuracy and overall reasoning performance. Our results highlight the need for improved reasoning abilities in dynamic video tasks, and position COVER as a new standard for evaluating multimodal logical reasoning.

\section*{Acknowledgments}

We would like to thank the anonymous reviewers for their valuable feedback. We thank Junshu Pan, Panzhong Lu, Fang Guo, Zijie Yang, Pai Liu, and other global collaborators for their valuable discussions and help. This work is funded by the National Natural Science Foundation of China Key Program (Grant No. 62336006), the Pioneer and “Leading Goose” R\&D Program of Zhejiang (Grant No. 2022SDXHDX0003), and the Research Program (Grant No. WU2023C020) of the Research Center for Industries of the Future, Westlake University.

\section*{Limitations}
COVER offers a novel benchmark for counterfactual video reasoning, but some limitations exist. First, while it focuses on video reasoning, its applicability to other multimodal tasks, such as image or text reasoning, remains unexplored. Second, COVER relies on sub-question decomposition, and automated methods may not always match human-designed questions, especially in complex scenarios. Finally, while we demonstrate COVER’s effectiveness on various models, further validation across different model architectures and real-world tasks is needed to assess its generalizability.

\section*{Ethical Considerations}
COVER is designed with ethical considerations in mind, aiming to enhance counterfactual reasoning in video understanding while ensuring fairness, transparency, and responsible AI development. We acknowledge the ongoing challenges in bias mitigation, fairness, and environmental sustainability and encourage the broader research community to collaborate in addressing these concerns. By establishing COVER as an open and structured evaluation benchmark, we aim to promote robust and ethical AI advancements in multimodal reasoning.

We ensured that the human annotators were compensated with fair remuneration, which exceeded the local minimum wage standards, reflecting the value of their work. Furthermore, we took steps to ensure that the annotation process did not pose any risks to their physical or mental well-being. The tasks were designed to be manageable, and we provided adequate support to ensure a safe and respectful working environment. 

In this study, AI was used solely for data augmentation and grammar/typo correction, with no involvement in generative or creative tasks. We carefully considered potential risks to ensure AI usage did not compromise the originality or transparency of the research.

\bibliography{custom}

\appendix

\section{Appendix}
\label{sec:appendix}

\subsection{Data Construction Details}
In this section, we present additional details on COVER construction, including information about the task splitting scores, annotation agreements, data augmentation prompts and process flow. 

We invited three expert annotators to independently score each benchmark task based on our two-dimensional quadrant framework (abstract vs. concrete and perception vs. cognition). Their scoring results in Table \ref{tab:score} demonstrates the strictness, consistency, and logical coherence of our task categorization, effectively preventing overlaps and ambiguity.

\begin{table*}[htbp]
\centering
\small
\begin{tabular}{lcccccccc}
\toprule
Task & $A_x$ & $A_y$ & $B_x$ & $B_y$ & $C_x$ & $C_y$ & Avg$_x$ & Avg$_y$ \\
\midrule
Counting & -3.2 & -3.4 & -3.1 & -3.6 & -3.3 & -3.7 & -3.2 & -3.57 \\
Color & -4.1 & -4.4 & -4.4 & -4.2 & -4.2 & -4.3 & -4.23 & -4.3 \\
Material & -3.8 & -3.3 & -3.9 & -3.2 & -4.0 & -3.4 & -3.9 & -3.3 \\
Size & -2.4 & -2.5 & -2.6 & -2.3 & -2.2 & -2.4 & -2.4 & -2.4 \\
Shape & -3.3 & -3.2 & -3.5 & -3.2 & -3.8 & -4.0 & -3.53 & -3.47 \\
Emotion & -2.4 & 4.0 & -2.5 & 3.5 & -2.4 & 3.1 & -2.43 & 3.53 \\
Location & -1.7 & -1.4 & -2.0 & -1.6 & -1.3 & -1.7 & -1.67 & -1.57 \\
Direction & -2.1 & -1.7 & -2.5 & -1.5 & -2.6 & -1.8 & -2.4 & -1.67 \\
Object Recognition & 3.0 & -3.0 & 2.4 & -2.0 & 1.2 & -2.3 & 2.2 & -2.43 \\
Action Recognition & 2.5 & -3.1 & 2.3 & -3.0 & 2.1 & -3.5 & 2.3 & -3.2 \\
Action Prediction & 3.9 & 2.4 & 3.8 & 2.5 & 3.2 & 2.2 & 3.63 & 2.37 \\
Procedure Understanding & 3.0 & 3.5 & 3.6 & 3.2 & 2.2 & 3.3 & 2.93 & 3.33 \\
Social Relation & 3.4 & 4.3 & 3.0 & 4.4 & 3.1 & 4.1 & 3.17 & 4.27 \\
\bottomrule
\end{tabular}
\caption{Annotator scoring table. Annotators A, B, and C provide ratings along two axes: the perceptual–cognitive dimension (x-axis, from $-5$ to $5$, where higher values indicate more cognitive tasks) and the concrete–abstract dimension (y-axis, from $-5$ to $5$, where higher values indicate more abstract tasks).}
\label{tab:score}
\end{table*}

\begin{figure*}[!htb]
\centering
  \includegraphics[width=0.75\textwidth]{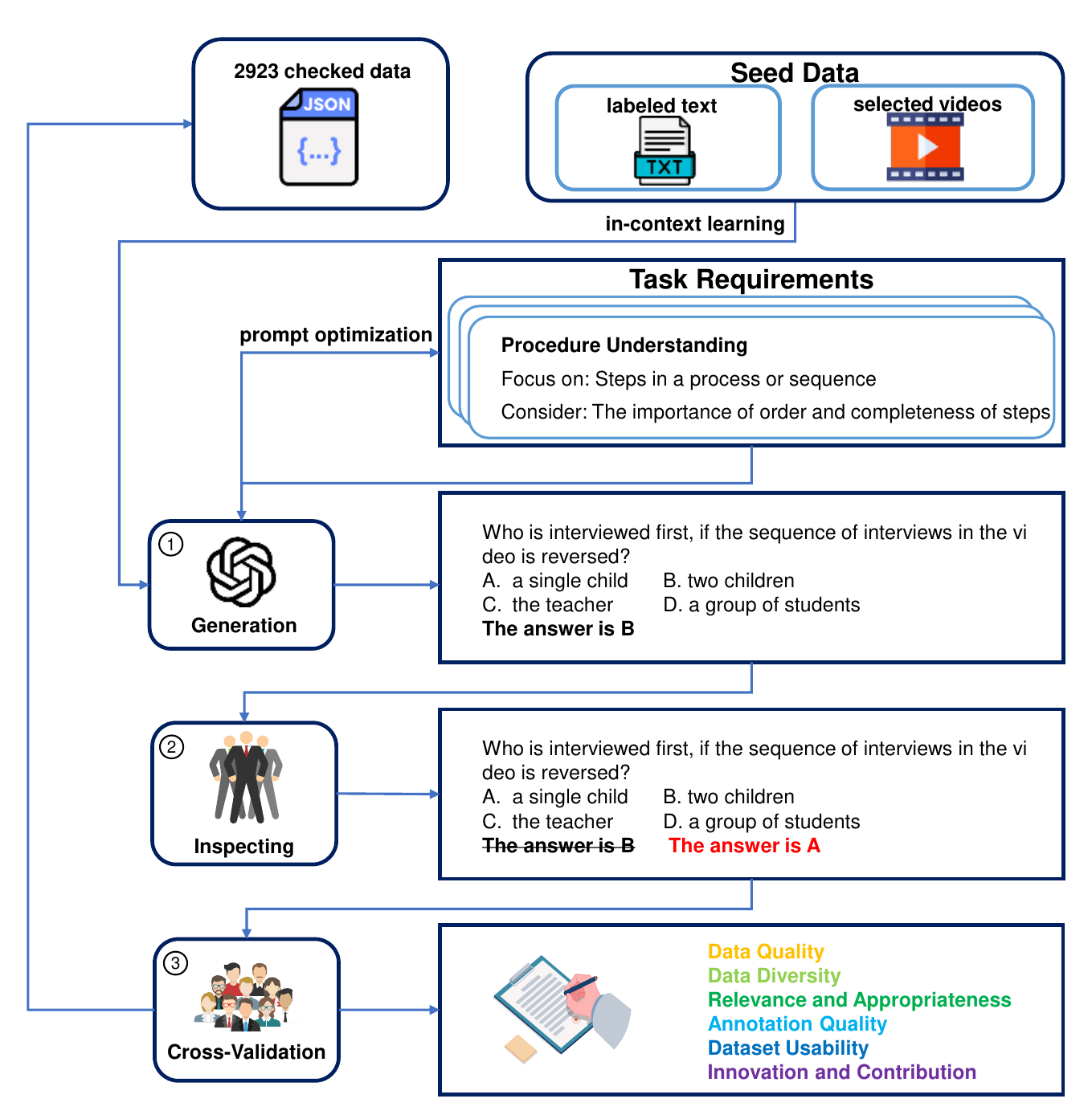}
  \caption{Flowchart depicting the data augmentation pipeline.}

  \label{fig:enlarge}
\end{figure*}

\begin{table}[h]
\centering
\small
\renewcommand{\arraystretch}{1.2}
\begin{tabular}{c|cccc}
\toprule
\textbf{Aspect} 
 & \textbf{A} & \textbf{B} & \textbf{C} & \textbf{Average} \\
\midrule
Data Quality         & 4 & 4 & 5 & 4.3 \\
\midrule
Data Diversity       & 5 & 4 & 5 & 4.7 \\
\midrule
Relevance            & 4 & 5 & 4 & 4.3 \\
\midrule
Annotation Quality   & 4 & 5 &  5& 4.7 \\
\midrule
Dataset Usability    & 4 & 4 & 4 & 4 \\
\midrule
Innovation           & 5 & 5 & 4 & 4.7 \\
\bottomrule
\end{tabular}
\caption{Cross-annotator validation on COVER. The table summarizes quality scores assigned by three annotators. A, B, and C denote randomly assigned codes for the assessment data, and \textit{Average} indicates the mean score across all entries.}
\label{tab:cross_val}
\end{table}
The annotators were recruited to evaluate COVER across multiple dimensions, with the resultant assessments systematically compiled in Table \ref{tab:cross_val}, ensuring comprehensive evaluation coverage. 
The methodological workflow for data augmentation is schematically outlined in Figure \ref{fig:enlarge}.

\begin{figure*}[!htb]
\centering
  \includegraphics[width=0.75\textwidth]{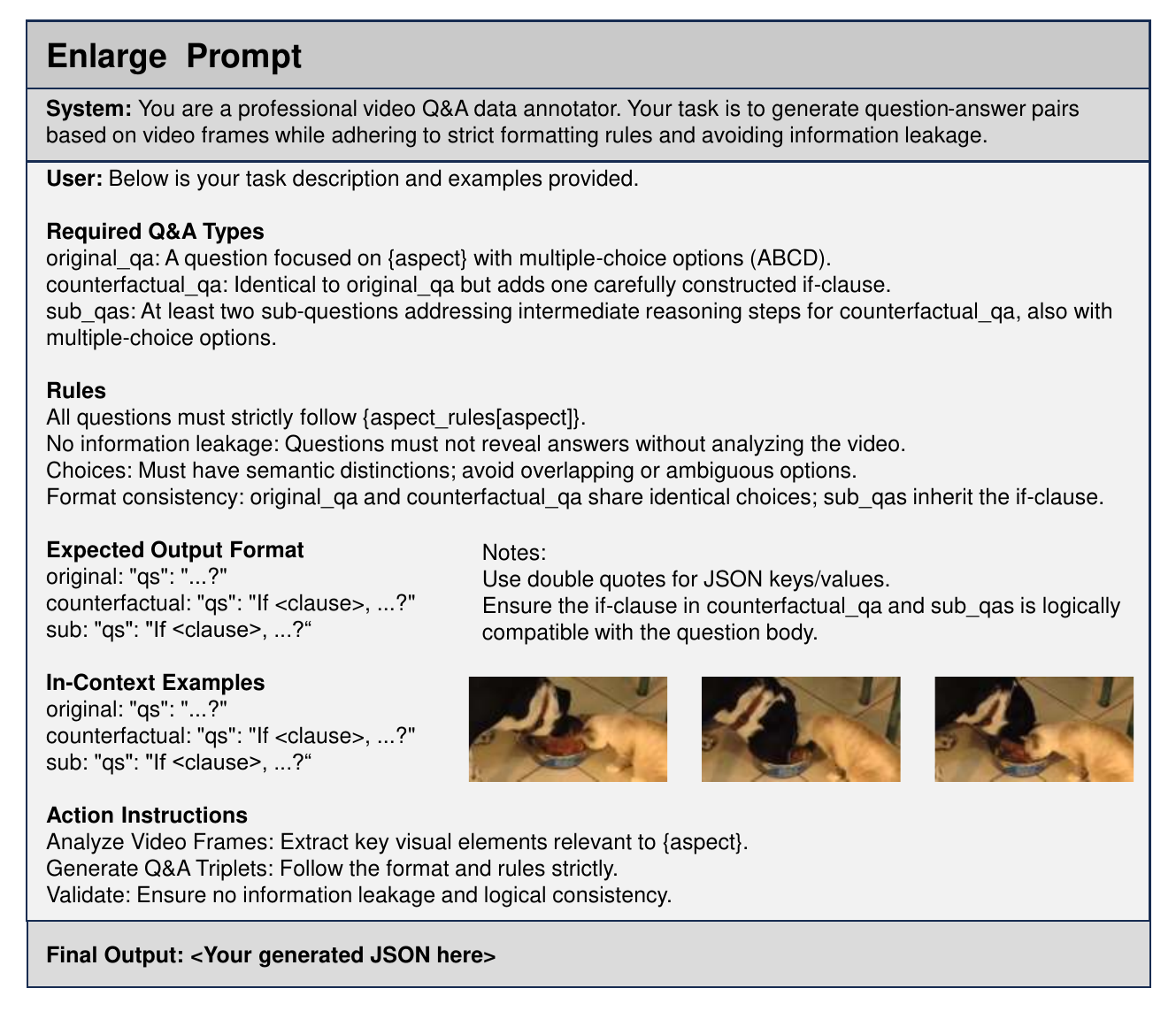}
\caption{Methodological framework for data augmentation using GPT-4o.}
  \label{fig:data_augmentation_prompts}
\end{figure*}


The schematic framework outlined in Figure \ref{fig:data_augmentation_prompts} delineates the methodology employed for contextual data augmentation, leveraging the generative capabilities of GPT-4o\cite{gpt4o} to construct domain-specific instructional prompts.


\subsection{Additional Results}

In this section, we present additional experiments on COVER. The comprehensive evaluation framework delineated in Table \ref{specific-eval-results} presents granular performance metrics across 13 meticulously defined tasks.

GPT-4o exhibited notable vulnerability in the Procedure Understanding task. While it attained a respectable raw accuracy of 78.17\%, its counterfactual accuracy plummeted to 28.97\%, representing a precipitous decline of 49.2\%. This substantial drop suggests that the performance of GPT-4o in understanding procedures may be overly reliant on surface-level features. Counterfactual perturbations, such as changes in conditions, can severely disrupt its reasoning capabilities, thereby highlighting a robustness limitation of the model when handling complex tasks.

Figure \ref{fig:acclinechart} (a) depicts the relationship between $ori_{acc}$ and $sub_{acc}$ across different models, with a purple regression line characterizing the functional correlation between mean $ori_{acc}$ and mean $sub_{acc}$. Figure \ref{fig:acclinechart} (b) demonstrates the association between $cf_{acc}$ and $sub_{acc}$ across different models, with a red regression line characterizing the functional correlation between mean $cf_{acc}$ and mean $sub_{acc}$. The bivariate correlation analysis delineated in Figure \ref{fig:acclinechart} demonstrates statistically significant covariation patterns (r = 0.836) between semantic comprehension and multi-step reasoning capabilities in MLLMs.

We conducted an additional ablation study to examine whether the observed trend where excessive visual information impairs reasoning accuracy holds consistently across both short and long videos. Our results are summarized in Table \ref{tab:short}, \ref{tab:long}. We observed a clear pattern across both short and long videos: model accuracy typically peaks within a moderate frame range (8--32 frames) and subsequently declines at the maximum setting (64 frames). This decline is particularly pronounced in tasks involving the original questions (ori) and sub questions (sub), suggesting that an excessive amount of visual input can indeed negatively impact model performance, regardless of video length.

\begin{table}[htbp]
    \centering
    \small
    \setlength{\tabcolsep}{4pt} 
    \begin{tabular}{c *{6}{r}}
        \toprule
        Frame & \multicolumn{3}{c}{InternVL2.5-4B} & \multicolumn{3}{c}{InternVL2.5-8B} \\
        \cmidrule(lr){2-4} \cmidrule(lr){5-7}
        & $ori_{acc}$ & $cf_{acc}$ & $sub_{acc}$ & $ori_{acc}$ & $cf_{acc}$ & $sub_{acc}$ \\
        \midrule
        2 & 69.09 & 45.94 & 60.33 & 68.72 & 56.53 & 61.14 \\
        4 & 70.81 & 46.18 & \textbf{60.91} & 68.97 & 56.90 & 60.68 \\
        8 & \textbf{71.31} & 43.97 & 60.62 & 69.83 & 56.28 & 61.14 \\
        16 & 70.81 & 44.83 & 59.86 & \textbf{70.07} & 56.40 & 61.20 \\
        32 & 70.69 & 43.84 & 59.57 & 69.21 & 56.90 & \textbf{61.26} \\
        64 & 69.95 & \textbf{46.55} & 59.63 & 69.09 & \textbf{57.27} & 60.62 \\
        \bottomrule
    \end{tabular}
    \caption{Performance of MLLMs with different numbers of sampled frames for short videos (1--64 frames).}
    \label{tab:short}
\end{table}

\begin{table}[htbp]
    \centering
    \small
    \setlength{\tabcolsep}{4pt} 
    \begin{tabular}{c *{6}{r}}
        \toprule
        Frame & \multicolumn{3}{c}{InternVL2.5-4B} & \multicolumn{3}{c}{InternVL2.5-8B} \\
        \cmidrule(lr){2-4} \cmidrule(lr){5-7}
        & $ori_{acc}$ & $cf_{acc}$ & $sub_{acc}$ & $ori_{acc}$ & $cf_{acc}$ & $sub_{acc}$ \\
        \midrule
        2 & 73.90 & \textbf{49.22} & 61.09 & 72.24 & \textbf{59.26} & 59.67 \\
        4 & 75.79 & 46.61 & 62.13 & 75.04 & 59.07 & 61.07 \\
        8 & 76.46 & 46.23 & 62.24 & 75.79 & 57.37 & 61.78 \\
        16 & \textbf{77.45} & 45.38 & \textbf{62.31} & 75.94 & 58.27 & \textbf{61.82} \\
        32 & 77.40 & 45.57 & 61.49 & 75.98 & 57.08 & 61.49 \\
        64 & 76.13 & 46.57 & 61.11 & \textbf{76.17} & 58.41 & 61.55 \\
        \bottomrule
    \end{tabular}
    \caption{Effect of different frame sampling strategies on MLLM performance for long videos (64--2000 frames).}
    \label{tab:long}
\end{table}

 Additionally, we evaluated test-time reasoning strategies on manually curated seed data using long-chain reasoning models in Table \ref{tab:gap_under_strategy}. Notably, models such as InternVL2.5-78B-CoT show significant improvement in bridging the cf–sub–ori gap, further supporting that reasoning-guided prompting (e.g., CoT) helps align sub-level and cf-level accuracy. These observations suggest a promising direction: larger and better-aligned models, when combined with explicit reasoning strategies, are more capable of maintaining coherence across perception, decomposition, and abstract reasoning tasks.

 \begin{table}[htbp]
\centering
\small
\begin{tabular}{lccc}
\toprule
Model & ori\_acc & cf\_acc & sub\_acc \\
\midrule
QVQ-72B-Preview & 69.33 & 59.33 & 58.76 \\
InternVL2.5-78B-CoT & 70.00 & 71.33 & 70.80 \\
\bottomrule
\end{tabular}
\caption{Variation in accuracy across different test-time reasoning strategies.}
\label{tab:gap_under_strategy}
\end{table}

\subsection{Sample Reaults on Test Time Long Reasoning Models}

\begin{figure*}[!htb]
\centering
  \includegraphics[width=0.85\textwidth]{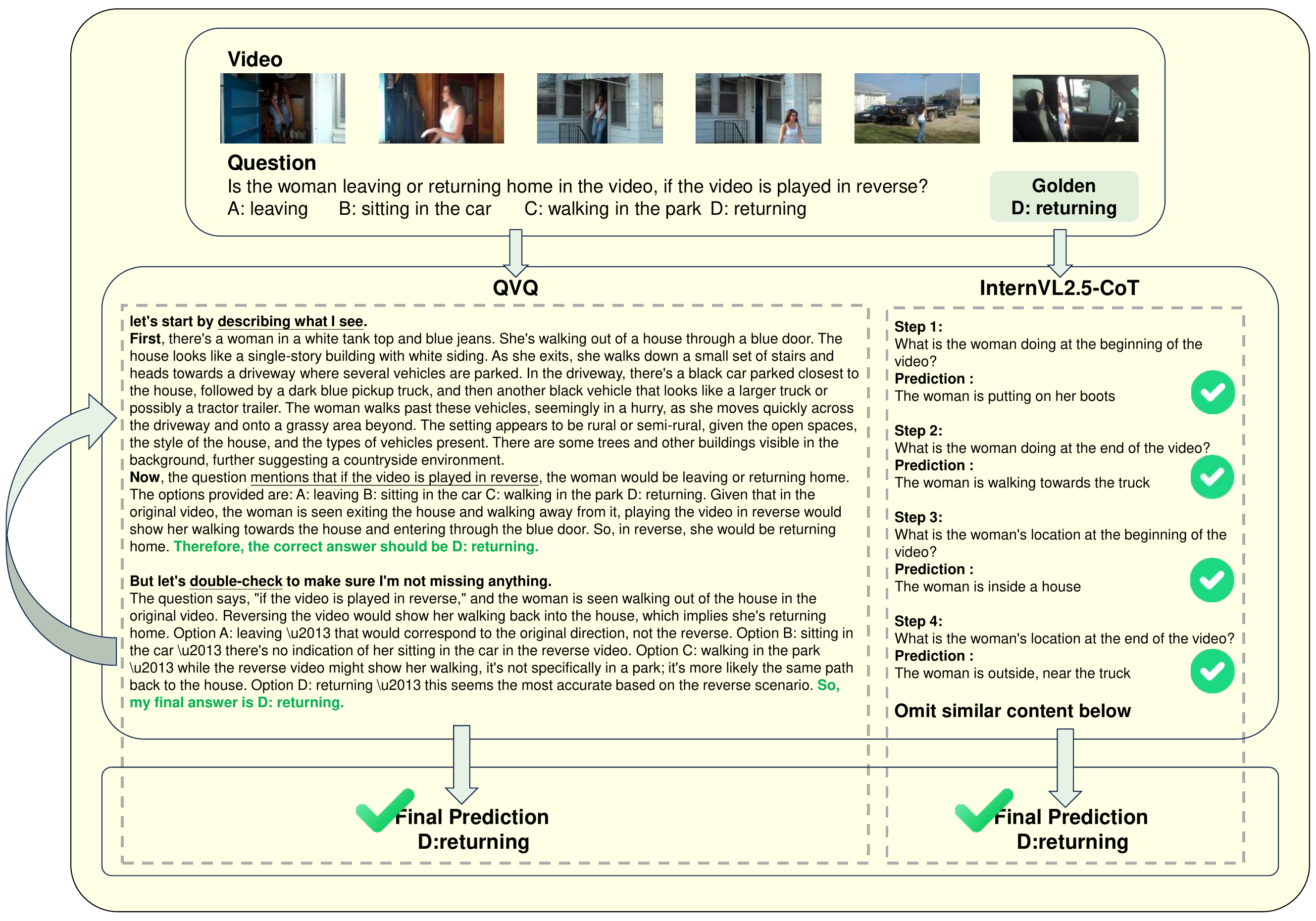}
  \caption{An example from the 150 seed samples. The top section shows the video input and corresponding counterfactual questions. The middle section compares two reasoning frameworks: the test-time long reasoning model QVQ and InternVL2.5-78B with CoT, with green marks indicating validated response components. The bottom section displays final model predictions, where green checkmarks indicate correct answers.}
  \label{fig:case3}
\end{figure*}

As illustrated in Figure~\ref{fig:case3}, the reasoning model QVQ-72B-Preview~\cite{qvq-72b-preview}, equipped with a built-in Chain-of-Thought (CoT) mechanism, exhibits human-aligned reasoning patterns. Its cognitive process integrates detailed scenario descriptions, systematic elimination of implausible options (e.g., excluding candidates A/B/C), and rigorous conclusion verification. In contrast, InternVL2.5-78B employs a CoT mechanism that presents answers in a bullet-point format without explanatory justification, reflecting weaker anthropomorphic reasoning characteristics.

However, the $cf_{acc}$ discrepancy in Table~\ref{tab:eval150_reasoning} (QVQ-72B-Preview: 59.33\% < InternVL2.5-78B: 71.33\%) suggests that contemporary reasoning models may rely more on memorization than on structured reasoning. InternVL2.5-78B's concise response paradigm appears to leverage rapid pattern recognition and information retrieval, leading to superior accuracy. While QVQ-72B-Preview's elaborate reasoning workflow better approximates human cognition, potential redundancies or logical inconsistencies may reduce answer precision.

Table~\ref{tab:eval150_reasoning} further indicates that InternVL2.5-78B achieves a substantial lead in the $sub_{acc}$ metric (70.80\%), significantly outperforming QVQ-72B-Preview (58.76\%) and Claude-3.7-sonnect (46.72\%). This performance hierarchy remains consistent across models when evaluated on the $ori_{acc}$ metric: InternVL2.5-78B (70.00\%) > QVQ-72B-Preview (69.33\%) > Claude-3.7-sonnect (46.00\%). Empirical evidence suggests a statistically significant positive correlation between reasoning capability ($sub_{acc}$) and comprehension ability ($ori_{acc}$). In addition, under the CoT paradigm, reasoning capability demonstrates stronger generalization, exhibiting a positive correlation with performance on human-annotated essential logical sub-problems, thereby reinforcing the intrinsic relationship between logical reasoning and generalizability.

\begin{figure*}[!htb]
\centering
  \includegraphics[width=1.0\textwidth]{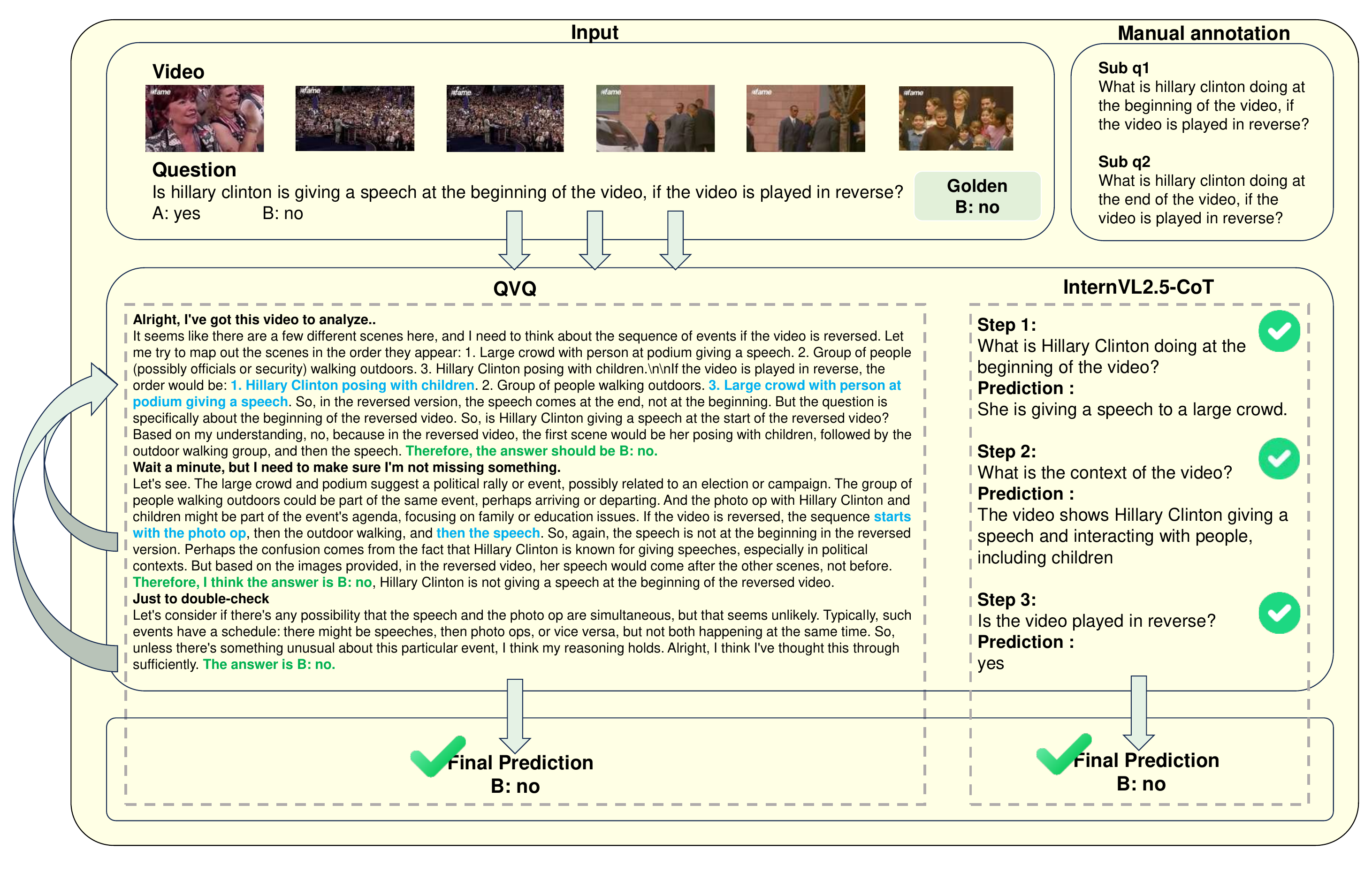}
  \caption{An example from the 150 seed samples. The top section presents the video input and corresponding counterfactual questions. The middle section compares QVQ and InternVL2.5-78B with CoT, using a dual-color annotation scheme: blue indicates conceptual alignment with manual sub-problem annotations, and green highlights validated response components. The bottom section shows the final model predictions, with green checkmarks indicating correct answers.}
  \label{fig:case4}
\end{figure*}

Moreover, the reasoning processes of models such as QVQ frequently generate sub-problem content that aligns with human-annotated data, which to some extent suggests that the inferential patterns of test-time long-reasoning models demonstrate closer correspondence with human cognitive intuition. For instance, in the Figure \ref{fig:case4} the analytical content regarding the opening and closing scenes of videos (highlighted in blue font) exhibits precise alignment with the manually curated sub-problems in the upper-right annotation (specifically addressing inquiries about video commencement and conclusion scenarios), thereby empirically validating this cognitive congruence.

\begin{table}
    \centering
    \small
    \begin{tabular}{lccc}
    \toprule
     & \makecell{\textbf{$ori_{acc}$}} & \makecell{\textbf{$cf_{acc}$}} & \makecell{\textbf{$sub_{acc}$}} \\
    \midrule
    QVQ-72B-Preview  &69.33 & 59.33 & 58.76 \\
    Claude-3.7-sonnect &46.00 & 59.33 & 46.72 \\
    \midrule
    InternVL2.5-78B &70.00 & 71.33 & 70.80 \\
    
    VILA1.5-13B &65.33 & 44.67 & 53.65 \\
    \bottomrule
    \end{tabular}
    \caption{Performance of different chain-of-thought (CoT) reasoning architectures on a manually annotated dataset of 150 samples. QVQ and Claude-3.5-Sonnet represent dedicated reasoning models, while the others apply CoT-based augmentation.}
    \label{tab:eval150_reasoning}
    \vspace{-2mm}
\end{table}

\begin{table*}
\centering
\begin{adjustbox}{max width=\linewidth}
\begin{tabular}{crccccccccccccc} 
\toprule
\multirow{3}{*}{\textbf{Model}} & \multicolumn{1}{c}{\multirow{3}{*}{\textbf{Type}}} & \multicolumn{13}{c}{\textbf{Task }}\\ 
\cmidrule{3-15}
                                 & \multicolumn{1}{c}{} 
                                 & \multicolumn{1}{c}{\multirow{1}{*}{\textit{Action}}} 
                                 & \multicolumn{1}{c}{\multirow{1}{*}{\textit{Procedure}}} 
                                 & \multicolumn{1}{c}{\textit{Social}}      
                                 & \multicolumn{1}{c}{\textit{Action}}   
                                 & \multicolumn{1}{c}{\textit{Object}}   
                                 & \multicolumn{1}{c}{\multirow{2}{*}{Color}} 
                                 & \multicolumn{1}{c}{\multirow{2}{*}{Counting}}
                                 & \multicolumn{1}{c}{\multirow{2}{*}{Direction}} 
                                 & \multicolumn{1}{c}{\multirow{2}{*}{Location}}
                                 & \multicolumn{1}{c}{\multirow{2}{*}{Material}} 
                                 & \multicolumn{1}{c}{\multirow{2}{*}{Shape}}
                                 & \multicolumn{1}{c}{\multirow{2}{*}{Size}} 
                                 & \multicolumn{1}{c}{\multirow{2}{*}{Emotion}}\\
                                 
                                 & \multicolumn{1}{c}{}                               
                                 & \multicolumn{1}{c}{\textit{Prediction}}                        
                                 & \multicolumn{1}{c}{\textit{Understanding}}                   
                                 & \multicolumn{1}{c}{\textit{Relation}} 
                                 & \multicolumn{1}{c}{\textit{Recognition}} 
                                 & \multicolumn{1}{c}{\textit{Recognition}}
                                 & \multicolumn{1}{c}{}                                        
                                 & \multicolumn{1}{c}{}
                                 & \multicolumn{1}{c}{}                                        
                                 & \multicolumn{1}{c}{}
                                 & \multicolumn{1}{c}{}                                        
                                 & \multicolumn{1}{c}{}
                                 & \multicolumn{1}{c}{}                                        
                                 & \multicolumn{1}{c}{}                          \\ 
\midrule
\multirow{3}{*}{GPT-4o} 
& $ori_{acc}$ 
& 65.20 & 78.17 & 69.00 & 74.87 & 74.87 & 92.23 & 75.25 & 50.88 & 70.59 & 79.12 & 72.00 & 52.88 & 65.01 \\
& $cf_{acc}$ 
& 41.41 & 28.97 & 56.33 & 44.65 & 42.67 & 37.86 & 40.59 & 33.33 & 42.86 & 59.34 & 58.00 & 29.81 & 55.65 \\
& $sub_{acc}$ 
& 51.85 & 22.82 & 52.43 & 69.54 & 67.09 & 51.94 & 47.52 & 56.90 & 48.96 & 58.08 & 55.56 & 36.08 & 63.97 \\
\midrule
\multirow{3}{*}{GPT-4o-mini} 
& $ori_{acc}$ 
& 50.22 & 72.22 & 63.32 & 78.61 & 74.08 & 84.47 & 70.30 & 52.63 & 57.98 & 71.43 & 62.00 & 56.73 & 65.56 \\
& $cf_{acc}$ 
& 44.05 & 51.19 & 62.01 & 56.42 & 52.36 & 26.21 & 39.60 & 36.84 & 59.66 & 52.75 & 53.00 & 17.31 & 58.26 \\
& $sub_{acc}$ 
& 53.16 & 19.44 & 58.85 & 68.03 & 64.82 & 38.35 & 38.12 & 53.88 & 47.72 & 53.89 & 53.44 & 28.85 & 65.96 \\
\midrule
\multirow{3}{*}{Claude-3.5-Sonnet} 
& $ori_{acc}$ 
& 43.61 & 63.10 & 63.32 & 66.31 & 73.82 & 79.61 & 68.32 & 48.25 & 52.10 & 63.74 & 66.34 & 45.19 & 66.94 \\
& $cf_{acc}$ 
& 39.21 & 33.33 & 38.86 & 43.85 & 40.84 & 36.89 & 19.80 & 35.96 & 40.34 & 37.36 & 40.59 & 18.27 & 39.81 \\
& $sub_{acc}$ 
& 46.19 & 15.87 & 46.68 & 62.54 & 60.93 & 39.81 & 24.26 & 48.28 & 48.55 & 46.11 & 37.70 & 34.13 & 56.81 \\
\midrule
\multirow{3}{*}{Gemini-1.5-Pro} 
& $ori_{acc}$ 
& 54.63 & 80.95 & 71.62 & 83.42 & 80.89 & 84.47 & 73.27 & 61.40 & 76.47 & 81.32 & 67.33 & 59.62 & 75.48 \\
& $cf_{acc}$ 
& 46.70 & 29.76 & 58.52 & 45.45 & 58.38 & 46.60 & 37.62 & 35.96 & 42.86 & 54.95 & 55.45 & 36.54 & 57.99 \\
& $sub_{acc}$ 
& 57.52 & 39.68 & 64.38 & 72.58 & 72.99 & 73.30 & 43.56 & 61.21 & 58.09 & 59.88 & 55.50 & 45.67 & 69.54 \\
\midrule
\multirow{3}{*}{Gemini-1.5-Flash} 
& $ori_{acc}$ 
& 53.74 & 85.32 & 70.74 & 82.62 & 81.41 & 82.52 & 70.30 & 57.02 & 70.59 & 79.12 & 69.31 & 68.27 & 72.04 \\
& $cf_{acc}$ 
& 45.81 & 34.92 & 56.33 & 49.20 & 49.48 & 41.75 & 37.62 & 33.33 & 41.18 & 64.84 & 53.47 & 25.96 & 58.26 \\
& $sub_{acc}$ 
& 61.87 & 32.94 & 63.94 & 73.28 & 69.60 & 46.60 & 43.07 & 62.93 & 54.77 & 62.87 & 55.50 & 37.98 & 71.36 \\
\midrule
\multirow{3}{*}{Gemini-2.0-Flash} 
& $ori_{acc}$   
& 60.35 & 86.90 & 74.24 & 87.97 & 80.10 & 90.29 & 69.31 & 64.04 & 78.99 & 81.32 & 70.30 & 66.35 & 75.90 \\
& $cf_{acc}$ 
& 42.29 & 36.51 & 51.97 & 44.12 & 51.31 & 20.39 & 39.60 & 31.58 & 37.82 & 57.14 & 56.44 & 31.73 & 57.71 \\
& $sub_{acc}$ 
& 60.78 & 35.52 & 59.51 & 73.16 & 72.49 & 69.90 & 59.41 & 65.95 & 53.53 & 58.08 & 58.64 & 42.31 & 66.84 \\
 \midrule
\multirow{3}{*}{InternVL2.5-78B} 
& $ori_{acc}$ 
& 67.84 & 75.00 & 75.55 & 79.68 & 82.20 & 94.17 & 82.18 & 52.63 & 76.47 & 76.92 & 83.17 & 69.23 & 76.86 \\
& $cf_{acc}$ 
& 43.61 & 76.19 & 57.21 & 65.51 & 61.78 & 87.38 & 37.62 & 47.37 & 75.63 & 61.54 & 57.43 & 39.43 & 56.20 \\
& $sub_{acc}$ 
& 62.09 & 44.64 & 67.70 & 76.90 & 62.28 & 79.13 & 69.80 & 66.38 & 58.09 & 62.28 & 59.69 & 50.48 & 70.07 \\
\midrule
\multirow{3}{*}{LLaVA-Video-72B}
& $ori_{acc}$ 
& 43.17 & 50.79 & 65.50 & 60.70 & 69.90 & 85.44 & 69.31 & 51.75 & 73.11 & 74.73 & 61.39 & 61.54 & 70.66 \\
& $cf_{acc}$ 
& 44.93 & 59.92 & 59.39 & 63.10 & 57.85 & 62.14 & 42.57 & 47.37 & 66.39 & 53.85 & 51.49 & 41.35 & 56.20 \\
& $sub_{acc}$ 
& 59.26 & 32.94 & 69.47 & 67.56 & 66.46 & 63.59 & 52.97 & 61.21 & 45.23 & 55.69 & 53.40 & 43.27 & 70.01 \\
 \midrule
\multirow{3}{*}{InternVL2.5-26B} 
& $ori_{acc}$ 
& 57.27 & 78.58 & 76.42 & 82.35 & 79.58 & 91.26 & 74.26 & 62.28 & 85.71 & 74.73 & 78.22 & 66.35 & 73.14 \\
& $cf_{acc}$ 
& 47.14 & 45.24 & 51.09 & 60.43 & 57.59 & 59.23 & 25.74 & 45.61 & 60.50 & 57.14 & 25.00 & 25.00 & 50.00 \\
& $sub_{acc}$ 
& 59.91 & 61.08 & 62.39 & 71.18 & 73.24 & 65.05 & 56.44 & 68.97 & 58.09 & 61/08 & 50.96 & 50.96 & 65.61 \\
 \midrule
\multirow{3}{*}{InternVL2.5-8B} 
& $ori_{acc}$ 
& 55.51 & 75.00 & 78.17 & 81.28 & 80.63 & 90.29 & 70.30 & 63.16 & 78.99 & 74.63 & 74.26 & 66.35 & 72.18 \\
& $cf_{acc}$ 
& 48.02 & 76.19 & 49.78 & 71.39 & 57.85 & 84.47 & 36.63 & 53.51 & 70.59 & 59.34 & 55.45 & 28.85 & 51.79 \\
& $sub_{acc}$ 
& 55.99 & 29.37 & 66.81 & 69.89 & 72.24 & 52.43 & 52.97 & 60.34 & 53.53 & 56.89 & 54.97 & 51.44 & 68.19 \\
 \midrule
\multirow{3}{*}{VideoLLama3-8B} 
& $ori_{acc}$ 
& 52.42 & 81.75 & 68.56 & 80.48 & 82.20 & 94.17 & 70.30 & 63.16 & 81.51 & 70.33 & 68.32 & 62.50 & 69.28 \\
& $cf_{acc}$ 
& 35.68 & 53.97 & 46.29 & 55.08 & 54.71 & 66.02 & 42.57 & 42.11 & 64.71 & 58.24 & 48.51 & 32.69 & 53.44 \\
& $sub_{acc}$ 
& 49.45 & 33.93 & 67.48 & 67.91 & 68.84 & 57.77 & 39.60 & 58.62 & 49.79 & 53.89 & 54.45 & 47.12 & 67.90 \\
 \midrule
\multirow{3}{*}{LLaVA-ov-7B} 
& $ori_{acc}$ 
& 48.90 & 48,81 & 66.81 & 60.43 & 65.45 & 86.41 & 63.37 & 44.74 & 63.03 & 72.53 & 61.39 & 63.46 & 68.60 \\
& $cf_{acc}$ 
& 43.61 & 64.29 & 45.85 & 55.35 & 50.79 & 59.22 & 42.57 & 45.61 & 52.94 & 60.44 & 57.43 & 30.77 & 52.75 \\
& $sub_{acc}$ 
& 50.11 & 30.36 & 63.94 & 62.78 & 60.68 & 54.85 & 42.08 & 53.45 & 45.64 & 50.90 & 48.69 & 50.96 & 64.73 \\
 \midrule
\multirow{3}{*}{LLaVA-Video-7B} 
& $ori_{acc}$ 
& 50.66 & 35.71 & 65.50 & 56.15 & 67.02 & 83.50 & 67.33 & 41.23 & 58.82 & 74.72 & 64.36 & 59.62 & 66.39 \\
& $cf_{acc}$ 
& 44.93 & 73.02 & 45.85 & 59.09 & 42.15 & 73.79 & 42.57 & 48.25 & 60.50 & 58.24 & 48.51 & 35.58 & 49.59 \\
& $sub_{acc}$ 
& 48.58 & 29.96 & 56.64 & 62.19 & 58.67 & 55.34 & 41.09 & 56.03 & 43.15 & 54.49 & 52.88 & 48.08 & 63.03 \\
 \midrule
\multirow{3}{*}{Qwen2-VL-7B} 
& $ori_{acc}$ 
& 44.49 & 84.12 & 67.25 & 84.22 & 80.10 & 88.35 & 70.29 & 57.89 & 73.94 & 74.73 & 65.34 & 69.23 & 67.49 \\
& $cf_{acc}$ 
& 42.29 & 58.73 & 45.41 & 44.92 & 41.88 & 56.31 & 21.78 & 42.11 & 58.82 & 60.44 & 45.54 & 33.65 & 49.72 \\
& $sub_{acc}$ 
& 53.37 & 30.16 & 63.72 & 67.33 & 66.71 & 43.20 & 42.08 & 59.48 & 51.87 & 52.69 & 51.83 & 51.44 & 65.02 \\
 \midrule
\multirow{3}{*}{VILA-U-7B} 
& $ori_{acc}$ 
& 45.37 & 73.02 & 54.59 & 66.31 & 59.95 & 81.55 & 61.39 & 54.39 & 71.43 & 47.25 & 47.52 & 47.11 & 59.50 \\
& $cf_{acc}$ 
& 22.47 & 53.17 & 42.36 & 44.39 & 39.53 & 45.63 & 45.54 & 23.68 & 43.70 & 35.16 & 35.64 & 36.54 & 33.88 \\
& $sub_{acc}$ 
& 38.78 & 34.33 & 44.03 & 52.16 & 57.04 & 41.26 & 22.28 & 39.22 & 40.66 & 35.93 & 42.93 & 42.31 & 55.34 \\
 \midrule
\multirow{3}{*}{VILA1.5-7B} 
& $ori_{acc}$ 
& 52.86 & 50.40 & 61.57 & 67.65 & 66.23 & 61.17 & 56.44 & 40.35 & 40.34 & 71.43 & 60.40 & 62.50 & 63.64 \\
& $cf_{acc}$ 
& 28.64 & 85.71 & 50.22 & 68.45 & 56.29 & 86.41 & 57.43 & 49.12 & 76.47 & 51.65 & 41.58 & 44.23 & 52.34 \\
& $sub_{acc}$ 
& 34.86 & 24.80 & 59.96 & 61.73 & 65.45 & 48.06 & 31.19 & 50.86 & 42.74 & 47.31 & 45.55 & 46.63 & 61.91 \\
\bottomrule
\end{tabular}
\end{adjustbox}
\caption{Overall performance of MLLMs on 13 tasks in COVER, including original accuracy, counterfactual accuracy, and sub-question accuracy.}
\label{specific-eval-results}
\end{table*}

\subsection{Examples of Sub-question Guidelines}

\begin{figure*}[!htb]
\centering
  \includegraphics[width=0.9\textwidth]{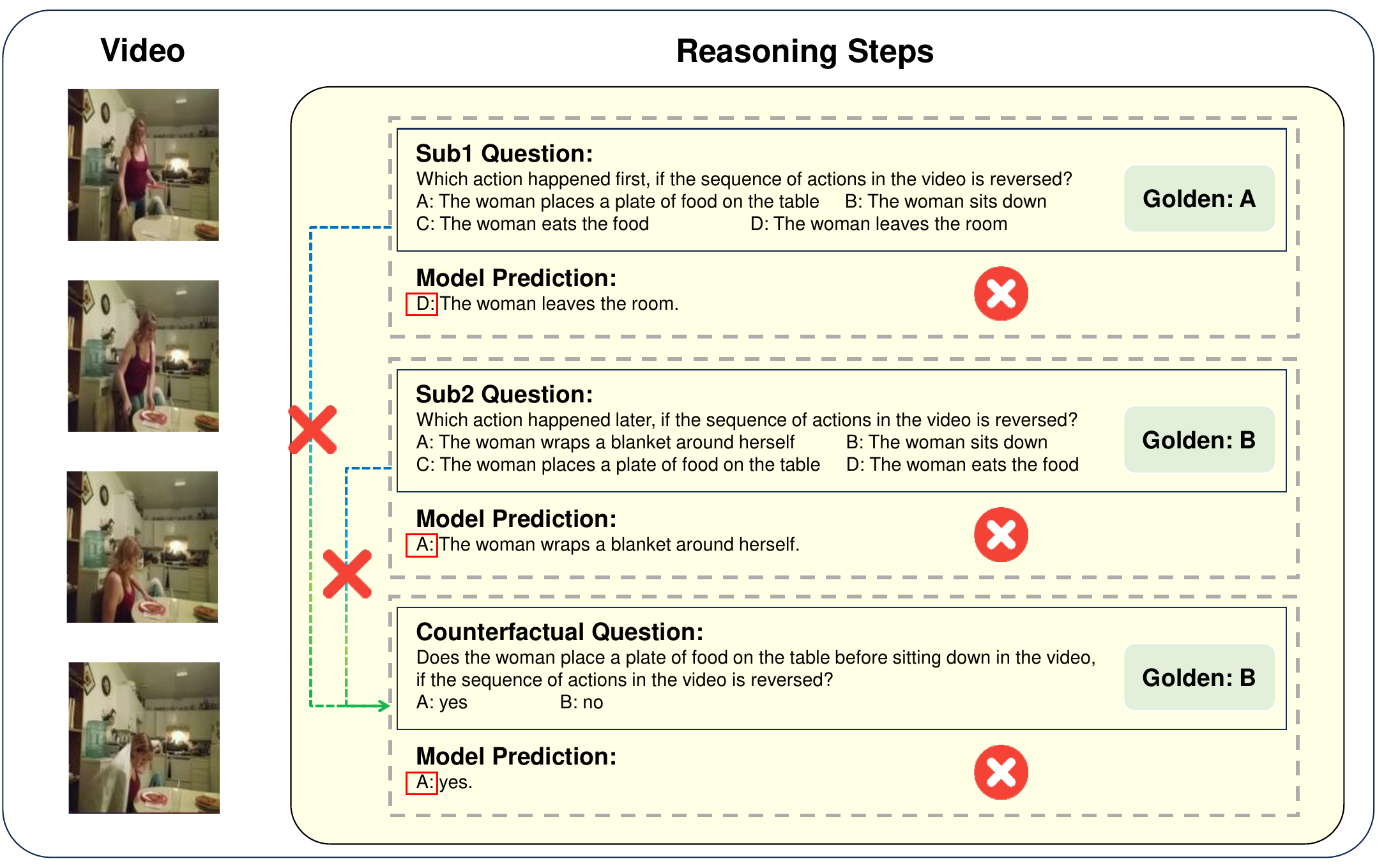}
  \caption{Example from \textbf{COVER}, showing a video accompanied by three related questions. The video is divided into four key action frames (left), with dashed lines indicating reasoning steps. Single-step prediction errors are marked with red crosses on the right, while sub-questions that do not support counterfactual reasoning are marked with red crosses on the left.}
  \label{fig:case1}
\end{figure*}

\begin{figure*}[!htb]
\centering
  \includegraphics[width=0.85\textwidth]{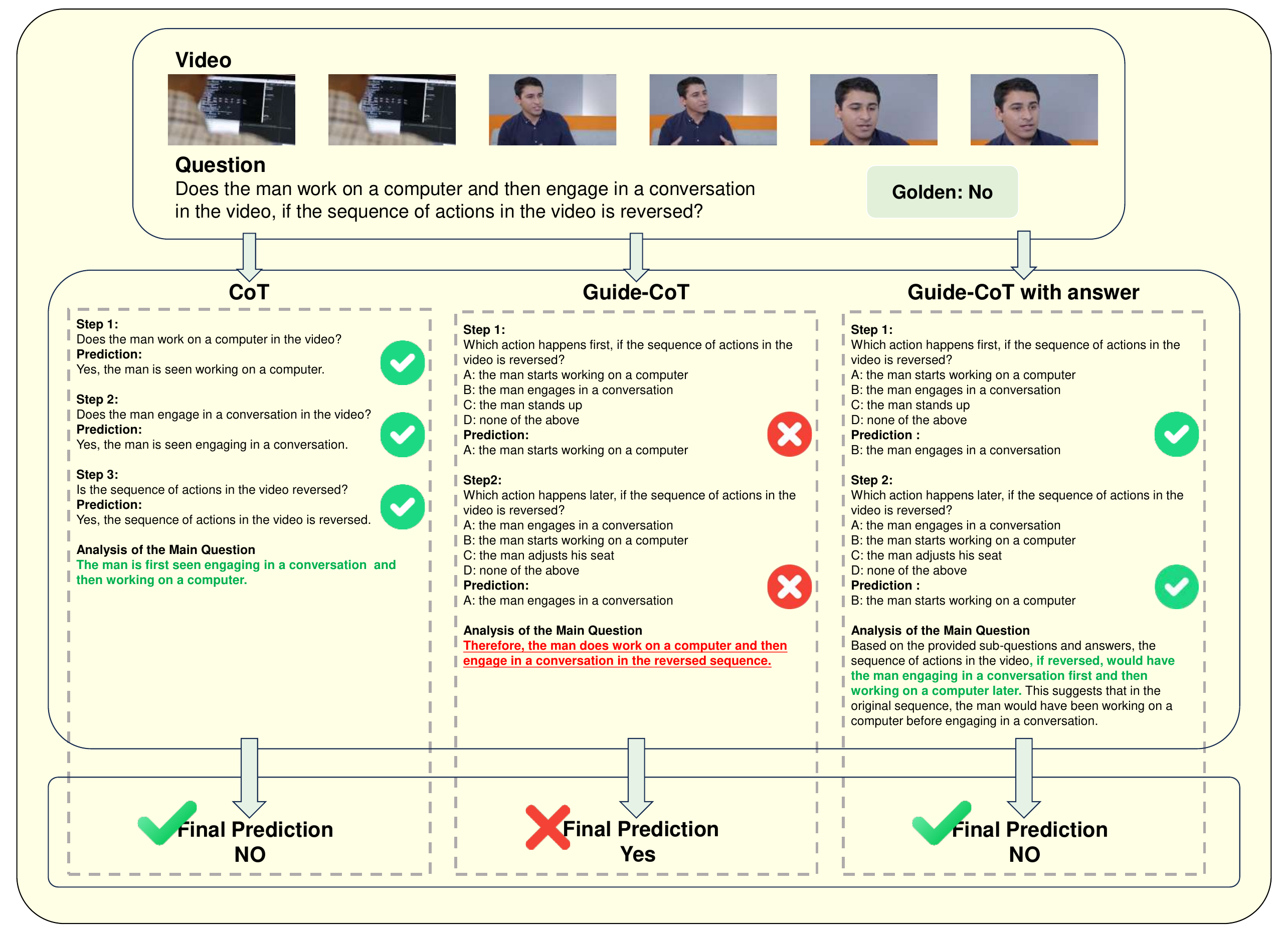}
  \caption{An example from \textbf{COVER}. The top section shows the video input and corresponding counterfactual questions. The middle section presents three reasoning processes—CoT, Guide-CoT, and Standard—where correct steps are marked with green checkmarks. In the analysis, correct reasoning paths are shown in green text, while incorrect ones are highlighted in red. The bottom section displays the final model predictions, with green checkmarks indicating correct answers and red crosses denoting errors.}
  \label{fig:case2}
\end{figure*}

Figure~\ref{fig:case1} illustrates how sub-question errors propagate to counterfactual question failures.
In Figure~\ref{fig:case2}, we observe that subtle errors in the reasoning process lead to reasoning failures, highlighting the model's sensitivity to the integrity of its reasoning steps.

\end{document}